\documentclass[preprint,12pt,authoryear]{elsarticle}

\usepackage{amsmath,amssymb,amsfonts}
\usepackage{algpseudocode,algorithm}
\usepackage{subcaption}
\usepackage{graphicx}
\usepackage{url}

\journal{Neural Networks}

\begin{document}

\sloppy

\begin{frontmatter}

\title{Pseudo-Quantized Actor-Critic Algorithm for Robustness to Noisy Temporal Difference Error} 

\author[nii,sokendai]{Taisuke Kobayashi} 
\ead{kobayashi@nii.ac.jp}
\ead[url]{https://prinlab.org/en/}

\affiliation[nii]{organization={Principles of Informatics Research Division, National Institute of Informatics (NII)},
            addressline={2-1-2 Hitotsubashi},
            city={Chiyoda-ku},
            postcode={101-8430},
            state={Tokyo},
            country={Japan}}

\affiliation[sokendai]{organization={Informatics Program, Graduate Institute for Advanced Studies (SOKENDAI)},
            addressline={2-1-2 Hitotsubashi},
            city={Chiyoda-ku},
            postcode={101-8430},
            state={Tokyo},
            country={Japan}}

\begin{abstract}
In reinforcement learning (RL), temporal difference (TD) errors are widely adopted for optimizing value and policy functions.
However, since the TD error is defined by a bootstrap method, its computation tends to be noisy and destabilize learning.
Heuristics to improve the accuracy of TD errors, such as target networks and ensemble models, have been introduced so far.
While these are essential approaches for the current deep RL algorithms, they cause side effects like increased computational cost and reduced learning efficiency.
Therefore, this paper revisits the TD learning algorithm based on control as inference, deriving a novel algorithm capable of robust learning against noisy TD errors.
First, the distribution model of optimality, a binary random variable, is represented by a sigmoid function.
Alongside forward and reverse Kullback-Leibler divergences, this new model derives a robust learning rule: when the sigmoid function saturates with a large TD error probably due to noise, the gradient vanishes, implicitly excluding it from learning.
Furthermore, the two divergences exhibit distinct gradient-vanishing characteristics.
Building on these analyses, the optimality is decomposed into multiple levels to achieve pseudo-quantization of TD errors, aiming for further noise reduction.
Additionally, a Jensen-Shannon divergence-based approach is approximately derived to inherit the characteristics of both divergences.
These benefits are verified through RL benchmarks, demonstrating stable learning even when heuristics are insufficient or rewards contain noise.
\end{abstract}

%


\end{frontmatter}

\section{Introduction}

Reinforcement learning (RL)~\citep{sutton2018reinforcement} can learn optimal policies through trial and error in (unknown) environments.
Particularly, deep RL (DRL)~\citep{mnih2015human}, when integrated with deep learning, has accelerated real-world practical applications by enabling flexible observation inputs including raw images to represent complex patterns of action:
such as, legged robot locomotion~\citep{radosavovic2024real}; autonomous driving~\citep{chen2021interpretable}; fine-tuning foundational models~\citep{ouyang2022training}; and finance~\citep{hambly2023recent}.
Improving DRL's learning performance enhances the capabilities of these applications and broadens the range of solvable problems.

The fundamental principle widely used in (D)RL algorithms is temporal difference (TD)~\citep{sutton2018reinforcement}.
This focuses on the temporal recursiveness inherent in the sum of rewards (i.e. return) and its expectation (i.e. value function), which serve as the RL objective function.
To learn the value function, TD learning employs a bootstrap method, namely the target is estimated by the value function (its approximator) itself.
Furthermore, the TD error defined as the difference between this target and the approximated value function can be regarded as the approximated advantage function~\citep{schulman2016high}.
Consequently, it is also widely utilized in policy gradient methods as advantage actor-critic algorithms~\citep{castro2008temporal,mnih2016asynchronous}.

However, since TD errors lack true targets, they become noisy as supervised learning signals.
Particularly in DRL, it is known that estimated targets fluctuate easily as noise due to the influence of nonlinear function approximation, causing learning instability~\citep{mnih2015human,van2016deep}.
Indeed, it has been demonstrated that increasing the frequency of value updates over policy updates improves estimation accuracy and yields superior learning performance~\citep{wang2025improving}, although the learning cost increases.
To stabilize learning, algorithms incorporate implementation techniques such as a target network that generates targets from other approximator with delayed parameters of the main approximator~\citep{mnih2015human}; and an ensemble model that utilizes statistic of estimates from multiple approximators as target~\citep{van2016deep}.
While these are heuristics, they have indeed stabilized learning and played a pivotal role in making DRL practical so far.

However, these heuristics carry side effects alongside the learning stability.
The target network is known to suffer from degraded sample efficiency due to delayed updates of targets~\citep{kim2019deepmellow}.
The ensemble model can exhibit discontinuous gradient fluctuations depending on the adopted statistic (i.e. min or max operator)~\citep{igel2024smooth}.
Furthermore, both heuristics require additional approximators, leading to increased memory footprint and computational cost during training.
In embedded systems like robots, where computational resources are constrained due to payload limitations, such burdens should not be ignored.
Note that improvements to mitigate the side effects of heuristics and other countermeasures for the noisy TD errors are summarized in the next section.

Against this background, this paper aims to theoretically discover a new algorithm capable of learning robustly even against noisy TD errors.
Specifically, a formulation based on \textit{control as inference}~\citep{levine2018reinforcement}, which allows for the theoretical introduction of nonlinearly transformed TD errors for gradient-based learning rules of value functions and policies~\citep{kobayashi2022optimistic,takahashi2025weber}.
That is, a new nonlinear transformation for the TD error that is robust to noise, i.e. saturation and quantization (or discretization)~\citep{casebeer2021enhancing,bernard2025cooperative}, is derived by utilizing (multiple) sigmoid functions in the probability model of the stochastic optimality variable.
Furthermore, considering that this nonlinear transformation is not uniquely determined, two types of gradients using Kullback-Leibler (KL) divergence based on the literature~\citep{kobayashi2022optimistic}.
To inherit their different characteristics, a new gradient-based learning rule that approximately uses Jensen-Shannon (JS) divergence is finally derived.

The performance of this proposed method, so-called pseudo-quantized actor-critic (PQAC), is evaluated through simulations using Mujoco~\citep{todorov2012mujoco}.
As a result, it is confirmed that PQAC can solve tasks more stably and efficiently than baselines.
Furthermore, PQAC can learn robustly even under conditions where the influences of the heuristics are weakened or where noise is injected into rewards.

\subsection{Related work}
\label{subsec:related}

\paragraph{Target network}
The target network is one of the popular tricks used to reduce fluctuations in targets required for calculating TD errors~\citep{mnih2015human}.
Typically, it shares the same structure as the main network model, and parameters (e.g. weights and biases) are gradually transferred from the main to target networks (e.g. copy every after the specified number of updates~\citep{mnih2015human} and interpolation between the main and target parameters~\citep{lillicrap2016continuous}).
This transfer delay contributes to stabilization, but conversely, it directly degrades sample efficiency in DRL.
Several methods have been proposed to mitigate this tradeoff, aiming to balance stability and efficiency.
For example, revising the parameter transfer rules to suppress the transfer of questionable parameters enables the generation of stable targets while improving transfer speed~\citep{kobayashi2021t}.
Lee and He~\citep{lee2019target} analyzed stabilization and efficiency when adding the regularization or constraint to the main network for suppressing divergence from the target network.
CAT-update~\citep{kobayashi2024consolidated}, which fuses these concepts, achieved further improvements, and will be utilized with the proposed method complementally in the simulations later.
Furthermore, taking a different approach, sharing all layers except the output layer between the main and target networks enables memory savings and suppression of the two divergences~\citep{vincent2025bridging}.

\paragraph{Ensemble model}
Instead of approximating the value function with a single network model, it is possible to improve the accuracy of value estimation by training multiple models in parallel and using statistic (e.g. mean) of their estimates as target~\citep{van2016deep}.
For example, popular DRL algorithms, SAC~\citep{haarnoja2018soft} and TD3~\citep{fujimoto2018addressing}, stabilize learning by training two independent networks and adopting the smaller of their estimates as the target for the TD error.
When limited to value updates, performance can be improved by setting the target as the mean of estimates from networks other than the one being updated~\citep{peer2021ensemble}, or by using the minimum estimate from randomly selected networks~\citep{chen2021randomized}.
Furthermore, as introduced in the target network paragraph, while sharing layers other than the output layer is possible to save memory, it is important to introduce random priors to ensure the estimates of networks are sufficiently diversified~\citep{osband2018randomized}.

\paragraph{Robust TD learning}
Learning rules with the TD error have been modified to alleviate the influences of inaccuracies in value estimation or noise in rewards.
This constitutes robustness against noise and is often referred to as robust TD learning.
Naively, the loss on the value function, i.e. the squared error for matching the classical TD learning rule, is replaced with a smooth L1 loss; and large gradients obtained after backpropagation are clamped~\citep{raffin2021stable,huang2022cleanrl}.
Both heuristics provide robustness to some extent when the target and estimated values diverge, but neither completely excludes erroneous estimates and data.
Meanwhile, the literature~\citep{cayci2023provably} has proposed the method that stop updates when the gradient norm exceeds a threshold, with their robustness demonstrated analytically.
However, designing this threshold remains a challenge in practice.
Although limited to the noise in rewards, a robust technique has been proposed for TD learning, which decomposes the reward-affected term and resets it to zero according to an analytically derived threshold~\citep{maity2025adversarially}.
As implied by the fact that many of the above techniques focus on gradients, DRL generally uses stochastic gradient descent to learn the value and policy functions.
Therefore, a strategy to make this process robust to noise is also conceivable.
This paper employs AdaTerm~\citep{ilboudo2023adaterm}, which is robust against gradient noise and outliers, as a complementary approach.
Note that integrating Monte Carlo method is a well-known representative technique for reducing biases arising from inacurate value estimation~\citep{schulman2016high,riquelme2019adaptive,gallici2025simplifying}.
However, it requires altering the data structure compared to standard TD learning, so it is omitted in this paper for prioritizing a fair comparison.

\section{Preliminaries}

\subsection{Reinforcement learning}

RL is to solve optimal control problems under Markov decision process (MDP) with the tuple $(\mathcal{S}, \mathcal{A}, \mathcal{R}, p_0, p_e)$~\citep{sutton2018reinforcement}.
Here, $\mathcal{S} \subseteq \mathbb{R}^{|\mathcal{S}|}$ and $\mathcal{A} \subseteq \mathbb{R}^{|\mathcal{A}|}$ denote the state and action spaces, respectively;
$\mathcal{R} \subseteq \mathbb{R}$ denotes the set of rewards;
and $p_0(s)$ and $p_e(s^\prime \mid s, a)$, with the current and next states $s, s^\prime \in \mathcal{S}$ and the action $a \in \mathcal{A}$, are the initial and state-transition probabilities.
An agent decides $a$ according to its state-dependent policy $\pi(a \mid s)$ (or a behavior policy $b(a \mid s)$ explicitly), then a faced (unknown) environment transitions from $s$ to $s^\prime$.
Each transition at time step $t$ is evaluated by a reward function $r_t = r(s, a, s^\prime)$, which represents the task to be solved.
By sequentially repeating the above step, the agent obtains the following return $R_t$.
\begin{align}
    R_t = \sum_{k=0}^\infty \gamma^k r_{t+k}
    \label{eq:return}
\end{align}
where $\gamma \in [0, 1)$ denotes the discount factor.

Under this MDP, RL aims to optimize $\pi(a \mid s)$ as follows:
\begin{align}
    \pi^*(\cdot \mid s) = \arg \max_{\pi} \mathbb{E}_{p_\tau}[R_t \mid s_t = s]
    \label{eq:prob_rl}
\end{align}
where $p_\tau$ denotes the probability of trajectory given by the chain of $p_e$ and $\pi$.
Note that $\mathbb{E}_{p_\tau}[R_t \mid s_t = s]$ is defined as the (state) value function $V^\pi(s)$ (or simply $V(s)$).
Actor-critic algorithms feature a structure where a critic learns the value function, and an actor improves the policy based on that evaluation.

\subsection{Control as inference}

The above optimal control problem has been revised as a kind of inference problem via \textit{control as inference}~\citep{levine2018reinforcement}.
To this end, a stochastic variable for optimality, $O = \{0, 1\}$, is additionally introduced to represent whether the future is optimal or not.
Its probability is modeled as the monotonically increasing function w.r.t. the value function, $f: \mathcal{R} \mapsto [0, 1]$.
\begin{align}
    \begin{split}
        p(O=1 \mid s) &= f(V(s))
        \\
        p(O=1 \mid s, a) &= f(Q(s,a))
    \end{split}
    \label{eq:prob_optim}
\end{align}
where $Q(s, a) = \mathbb{E}_{p_\tau}[R_t \mid s_t = s, a_t = a]$ denotes the action value function.
This can be approximated as $r + \gamma V(s^\prime)$ according to Bellman equation and Monte Carlo approximation.

In the literature~\citep{kobayashi2022optimistic}, the optimal and non-optimal policies are given through Bayes theorem as follows:
\begin{align}
    &\pi(a \mid s; O)
    = \cfrac{p(O \mid s, a) b(a \mid s)}{p(O \mid s)}
    \label{eq:policy_bayes_optim}
\end{align}
With this, the optimization problems of the value and policy functions (with parameters $\theta$ and $\phi$, respecitvely) are newly formulated.
\begin{align}
    &\min_\theta \mathbb{E}_{p_e, b}[d(p(O \mid s, a) \mid p(O \mid s))]
    \label{eq:problem_val} \\
    &\min_\phi \mathbb{E}_{p_e}[d(\pi(a \mid s; O=1) \mid \pi(a \mid s)) - d(\pi(a \mid s; O=0) \mid \pi(a \mid s))]
    \label{eq:problem_pol}
\end{align}
where $d(p \mid q)$ denotes the divergence between two probabilities, $p,q$.

When adopting exponential function as $f$ and reverse KL divergence as $d$, the basic (advantage) actor-critic algorithm is obtained.
That is, the gradients w.r.t. $\theta$ and $\phi$ are approximately derived as follows:
\begin{align}
    g_\theta &= \mathbb{E}_{p_e, b}\left[- \delta \nabla_\theta V(s) \right]
    \label{eq:grad_val_base} \\
    g_\phi &= \mathbb{E}_{p_e, \pi}\left[- \delta \nabla_\phi \ln \pi(a \mid s) \right]
    \label{eq:grad_pol_base}
\end{align}
where $\delta$ denotes the TD error, defined as:
\begin{align}
    \delta = r + \gamma V(s^\prime) - V(s)
    \label{eq:td_err}
\end{align}
Note that the probability distributions for calculating the expected operation differ between $b$ and $\pi$, but this can be addressed using importance sampling or the method that ensures $b \simeq \pi$~\citep{kobayashi2024revisiting}.

As a remark, by adopting ``forward'' KL divergence as $d$, the weight term of both gradients, $\delta$, is nonlinearly transformed to $\exp(\delta) - 1$ with optimism.
In addition, defining $f$ based on a origin-symmetric function w.r.t. the exponential function yields pessimistic update rules.

\subsection{Stabilization tricks for TD learning}

Focusing on the gradient of $V$ in eq.~\eqref{eq:grad_val_base}, we can interpret this as supervised learning where $r + \gamma V(s^\prime)$ is given as the target.
However, since the calculation of the target includes the estimate to be optimized, $V(s^\prime) = V^\prime$, this corresponds to a bootstrap method.
As a result, learning can easily become unstable due to estimation errors in $V^\prime$, not only when rewards contain noise.
This instability also affects policy improvement, necessitating tricks to reduce the estimation error of $V^\prime$.

As introduced in Section~\ref{subsec:related}, representative tricks include the target network and the ensemble model.
First, for the main network with parameters $\theta$, the target network has the same structure with different parameters $\bar{\theta}$, which are initialized as $\theta$.
When updating $\theta$ (and $\phi$), the target network is frozen to compute $V^\prime = V(s^\prime; \bar{\theta})$ stably.
Afterwards, $\bar{\theta}$ is updated towards $\theta$ for estimating $V^\prime$ accurately.
\begin{align}
    \bar{\theta} \gets g(\bar{\theta}, \theta)
\end{align}
where $g$ is the update rule, where $\bar{\theta} \to \theta$ holds eventually.
In this paper, CAT-soft update~\citep{kobayashi2024consolidated} is employed as mentioned before.
Note that the target network is sometimes applied to $\phi$, the policy model.

Second is the ensemble model.
That is, $\{\theta_k\}_{k=1}^K$, with $K$ the number of models, are initialized independently and trained in the same way.
Then, the target is computed with the specified statistic (the median in this paper, following the literature~\citep{kobayashi2023reward}), $m: \mathcal{R}^K \mapsto \mathcal{R}$, as follows:
\begin{align}
    V^\prime = m(V(s^\prime; \theta_1), \ldots, V(s^\prime; \theta_K))
\end{align}
The estimation inaccuracy of each model can be compensated with each other, stabilizing TD learning.

These two tricks can be employed simultaneously, and the implementation in this paper does so.
While the combination improves the accuracy of value estimation synergistically, it requires $2K$ models and is costly.
Furthermore, while stability would improve, there is a risk that sample efficiency might decrease.
Therefore, the introduction of these tricks should be minimized, and then, the learning algorithm should robustly work even with noisy TD errors.

\section{Pseudo-saturation through control as inference}

\subsection{Probability model for optimality}

As mentioned before, the exponential function is basically adopted to model the probability distribution of optimality.
While this is an excellent model in that its derived results match the conventional RL algorithms (i.e. TD learning and policy gradient), caution is required when using the exponential function as a probability mass function because it can output exceeding one.
We must estimate the utopia point of the value function and input the difference relative to it.
If this estimation is erroneous, it fails to satisfy the definition as a probability mass function.
The derivation in eqs.~\eqref{eq:grad_val_base} and~\eqref{eq:grad_pol_base} implicitly ignores this issue by employing an approximation.

\begin{figure}[tb]
    \centering
    \includegraphics[keepaspectratio=true,width=0.96\linewidth]{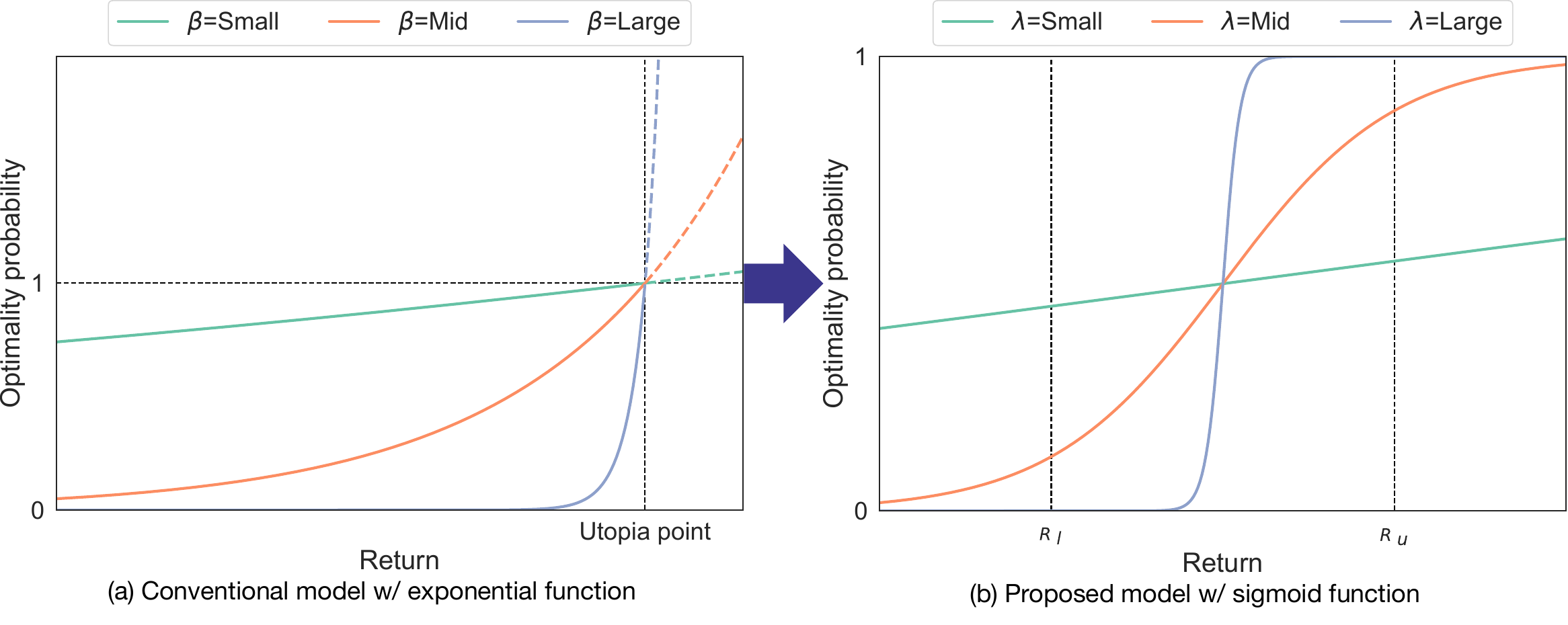}
    \caption{Replacement of the model for optimality probability:
    (a) the model with exponential function is usually employed due to its simplicity;
    (b) to always satisfy the definition of probability, a new model with sigmoid function is introduced in this work.
    }
    \label{fig:optimality_replace}
\end{figure}

With this in mind, this paper proposes the use of the sigmoid function $\sigma: \mathbb{R} \mapsto (0, 1)$ as a new model.
Specifically, eq.~\eqref{eq:prob_optim} is concretized as follows (also see Fig.~\ref{fig:optimality_replace}):
\begin{align}
    \begin{split}
        p(O=1 \mid s) &= \sigma(\lambda_O (V(s) - \mu_O)) = \cfrac{1}{1 + \exp(- \lambda_O (V(s) - \mu_O))}
        \\
        p(O=1 \mid s, a) &= \sigma(\lambda_O (Q(s,a) - \mu_O)) = \cfrac{1}{1 + \exp(- \lambda_O (Q(s,a) - \mu_O))}
    \end{split}
    \label{eq:prob_optim_prop}
\end{align}
where $\mu_O = (R_u + R_l) / 2$ and $\lambda_O^{-1} = (R_u - R_l) / (2 \lambda)$ with the upper and lower bounds of return, $R_u, R_l$, and $\lambda > 0$ the sharpness hyperparameter.
Note that $R_u$ and $R_l$ should be estimated appropriately, but even if their estimation is incorrect, the definition of the probability mass function is always satisfied due to $\sigma$, so estimation error is permissible and no approximation is needed to eliminate them.

Here, three characteristics of $\sigma$ that will be needed later are introduced.
First, $1-\sigma(x)$ with $x$ the input, which is necessary for $p(O=0)$, is simplified as follows:
\begin{align}
    1-\sigma(x) = \sigma(-x)
    \label{eq:sigma_neg}
\end{align}
Second, the derivative of $\sigma$ w.r.t. $x$ is well known to be with its own output.
\begin{align}
    \frac{d\sigma(x)}{dx} = \sigma(x)(1 - \sigma(x))
    \label{eq:sigma_grad}
\end{align}
Finally, the natural logarithm of $\sigma$ can be expressed using a softplus function, $\mathrm{sp}: \mathbb{R} \mapsto \mathbb{R}_+$.
\begin{align}
    \ln \sigma(x) = - \mathrm{sp}(-x) = x - \mathrm{sp}(x)
    \label{eq:sigma_log}
\end{align}

\subsection{Derivation with forward/reverse KL divergence}

Using the above model of optimality probability, the gradients for updating the value function and policy are analytically derived.
As a first step, the case where the forward or reverse KL divergence is adopted as $d$ (the divergence measure between distributions) is considered since they have been employed in previous studies~\citep{kobayashi2022optimistic}.
Note that the other study derived further properties by considering the inverse of optimality probability, but that is unnecessary for the symmetric sigmoid function due to the same result as the normal case.

\subsubsection{Case 1: reverse KL divergence}

With eq.~\eqref{eq:prob_optim_prop}, the reverse KL divergence is first introduced to $d$ in eqs.~\eqref{eq:problem_val} and~\eqref{eq:problem_pol} as follows:
\begin{align}
    &\min_\theta \mathbb{E}_{p_e, b}[\mathrm{KL}(p(O \mid s) \mid p(O \mid s, a))]
    \nonumber \\
    =& \min_\theta \mathbb{E}_{p_e, b}[ \sigma_V (\ln \sigma_V - \ln \sigma_Q) + \bar{\sigma}_V (\ln \bar{\sigma}_V - \ln \bar{\sigma}_Q) ]
    \label{eq:problem_val_rkl} \\
    &\min_\phi \mathbb{E}_{p_e}[\mathrm{KL}(\pi(a \mid s) \mid \pi(a \mid s; O=1)) - \mathrm{KL}(\pi(a \mid s) \mid \pi(a \mid s; O=0))]
    \nonumber \\
    =& \min_\phi \mathbb{E}_{p_e,\pi}[\ln \pi(a \mid s; O=0) - \ln \pi(a \mid s; O=1)]
    \label{eq:problem_pol_rkl}
\end{align}
where $\sigma_x = \sigma(\lambda_O (x - \mu_O))$ and $\bar{\sigma}_x = 1 - \sigma_x$ with $x = \{V,Q\}$ for simplicity.

Analytically calculating the gradient for each objective function to be minimized w.r.t. its corresponding parameter (i.e. $\theta$ or $\phi$) yields the following results.
\begin{align}
    g_\theta^\mathrm{RKL} &= \mathbb{E}_{p_e, b}[\nabla_\theta V(s) \lambda_O \sigma_V \bar{\sigma}_V (\ln \sigma_V - \ln \sigma_Q - \ln \bar{\sigma}_V + \ln \bar{\sigma}_Q)
    \nonumber \\
    &\quad\quad\quad\quad + \nabla_\theta \sigma_V \sigma_V / \sigma_V - \nabla_\theta \sigma_V \bar{\sigma}_V / \bar{\sigma}_V]
    \nonumber \\
    &= \mathbb{E}_{p_e, b}[\nabla_\theta V(s) \lambda_O \sigma_V \bar{\sigma}_V \{ \lambda_O (V(s) - \mu_O) - \mathrm{sp}(\lambda_O (V(s) - \mu_O))
    \nonumber \\
    &\quad\quad\quad\quad\quad\quad\quad\quad\quad\quad - \lambda_O (Q(s,a) - \mu_O) + \mathrm{sp}(\lambda_O (Q(s,a) - \mu_O))
    \nonumber \\
    &\quad\quad\quad\quad\quad\quad\quad\quad\quad\quad + \mathrm{sp}(\lambda_O (V(s) - \mu_O)) - \mathrm{sp}(\lambda_O (Q(s,a) - \mu_O))\}]
    \nonumber \\
    &= \mathbb{E}_{p_e, b}[- \nabla_\theta V(s) \lambda_O^2 \sigma_V \bar{\sigma}_V (Q(s,a) - V(s))]
    \label{eq:grad_val_rkl_raw}
\end{align}
\begin{align}
    g_\phi^\mathrm{RKL} &= \mathbb{E}_{p_e,\pi}[\nabla_\phi \ln \pi(a \mid s) \{\ln \pi(a \mid s; O=0) - \ln \pi(a \mid s; O=1)\}]
    \nonumber \\
    &= \mathbb{E}_{p_e,\pi}[\nabla_\phi \ln \pi(a \mid s) \{\ln \bar{\sigma}_Q b(a \mid s) - \ln \bar{\sigma}_V - \ln \sigma_Q b(a \mid s) + \ln \sigma_V\}]
    \nonumber \\
    &= \mathbb{E}_{p_e,\pi}[\nabla_\phi \ln \pi(a \mid s) \{- \mathrm{sp}(\lambda_O (Q(s,a) - \mu_O)) + \mathrm{sp}(\lambda_O (V(s) - \mu_O))
    \nonumber \\
    &\quad\quad\quad\quad\quad\quad\quad\quad\quad\quad - \lambda_O (Q(s,a) - \mu_O) + \mathrm{sp}(\lambda_O (Q(s,a) - \mu_O))
    \nonumber \\
    &\quad\quad\quad\quad\quad\quad\quad\quad\quad\quad + \lambda_O (V(s) - \mu_O) - \mathrm{sp}(\lambda_O (V(s) - \mu_O))\}]
    \nonumber \\
    &= \mathbb{E}_{p_e,\pi}[- \nabla_\phi \ln \pi(a \mid s) \lambda_O (Q(s,a) - V(s))]
    \label{eq:grad_pol_rkl_raw}
\end{align}
Note that eqs.~\eqref{eq:sigma_neg}--\eqref{eq:sigma_log} are utilized in the above transformations.

\subsubsection{Case 2: forward KL divergence}

Similarly, the forward KL divergence is introduced to $d$ in eqs.~\eqref{eq:problem_val} and~\eqref{eq:problem_pol} as follows:
\begin{align}
    &\min_\theta \mathbb{E}_{p_e, b}[\mathrm{KL}(p(O \mid s, a) \mid p(O \mid s))]
    \nonumber \\
    =& \min_\theta \mathbb{E}_{p_e, b}[ - \sigma_Q \ln \sigma_V - \bar{\sigma}_Q \ln \bar{\sigma}_V ]
    \label{eq:problem_val_fkl} \\
    &\min_\phi \mathbb{E}_{p_e}[\mathrm{KL}(\pi(a \mid s; O=1) \mid \pi(a \mid s)) - \mathrm{KL}(\pi(a \mid s; O=0) \mid \pi(a \mid s))]
    \nonumber \\
    =& \min_\phi \mathbb{E}_{p_e,b}\left[ \ln \pi(a \mid s) \left( \frac{\bar{\sigma}_Q}{\bar{\sigma}_V} - \frac{\sigma_Q}{\sigma_V} \right) \right ]
    \label{eq:problem_pol_fkl}
\end{align}
where the terms unrelated to $\theta$ or $\phi$ were excluded.

The gradients for them are analytically derived as well.
\begin{align}
    g_\theta^\mathrm{FKL} &= \mathbb{E}_{p_e, b}[ - \nabla_\theta V(s) \lambda_O \sigma_Q \bar{\sigma}_V + \nabla_\theta V(s) \lambda_O \bar{\sigma}_Q \sigma_V ]
    \nonumber \\
    &= \mathbb{E}_{p_e, b}[ - \nabla_\theta V(s) \lambda_O (\sigma_Q - \sigma_V) ]
    \label{eq:grad_val_fkl_raw}
\end{align}
\begin{align}
    g_\phi^\mathrm{FKL} &= \mathbb{E}_{p_e,b}\left[ \nabla_\phi \ln \pi(a \mid s) \left( \frac{\bar{\sigma}_Q}{\bar{\sigma}_V} - \frac{\sigma_Q}{\sigma_V} \right) \right ]
    \nonumber \\
    &= \mathbb{E}_{p_e,b}\left[ - \nabla_\phi \ln \pi(a \mid s) \frac{\sigma_Q - \sigma_V}{\sigma_V \bar{\sigma}_V} \right ]
    \label{eq:grad_pol_fkl_raw}
\end{align}

\subsubsection{Alignment of gradients}

The derived gradients are seen to possess a similar structure.
However, this differs from the conventional structure where the gradients of the value function and the log-policy are multiplied by identical weights (i.e. the TD error).
Furthermore, eq.~\eqref{eq:grad_pol_fkl_raw} is prone to be numerically unstable because its denominator can asymptotically approach zero.
Therefore, to align these gradients, $g_\phi^\mathrm{RKL,FKL}$ and $g_\theta^\mathrm{RKL,FKL}$ are multiplied by $\sigma_V \bar{\sigma}_V$ (which is proportional to the Fisher information of $p(O|V)$) and $1/\lambda_O$ (which can be considered a constant), respectively.
As a result, the following gradients are finally obtained.
\begin{align}
    g_\theta^\mathrm{RKL} &\propto \mathbb{E}_{p_e, b}[- \nabla_\theta V(s) \lambda_O \sigma_V \bar{\sigma}_V (Q(s,a) - V(s))]
    \label{eq:grad_val_rkl} \\
    g_\phi^\mathrm{RKL} &\propto \mathbb{E}_{p_e,\pi}[- \nabla_\phi \ln \pi(a \mid s) \lambda_O \sigma_V \bar{\sigma}_V (Q(s,a) - V(s))]
    \label{eq:grad_pol_rkl} \\
    g_\theta^\mathrm{FKL} &\propto \mathbb{E}_{p_e, b}[ - \nabla_\theta V(s) (\sigma_Q - \sigma_V) ]
    \label{eq:grad_val_fkl} \\
    g_\phi^\mathrm{FKL} &\propto \mathbb{E}_{p_e,b}[ - \nabla_\phi \ln \pi(a \mid s) (\sigma_Q - \sigma_V) ]
    \label{eq:grad_pol_fkl}
\end{align}

Here, by regarding the weight applied to each gradient as the nonlinearly transformed TD error, it for each divergence can be defined as follows:
\begin{align}
    \delta^\mathrm{RKL}(\delta) &= \lambda_O \sigma_V \bar{\sigma}_V \delta
    \label{eq:td_err_rkl} \\
    \delta^\mathrm{FKL}(\delta) &= \sigma_{\delta + V} - \sigma_V
    \label{eq:td_err_fkl}
\end{align}

\subsection{Analysis}

\begin{figure}[tb]
    \centering
    \includegraphics[keepaspectratio=true,width=0.96\linewidth]{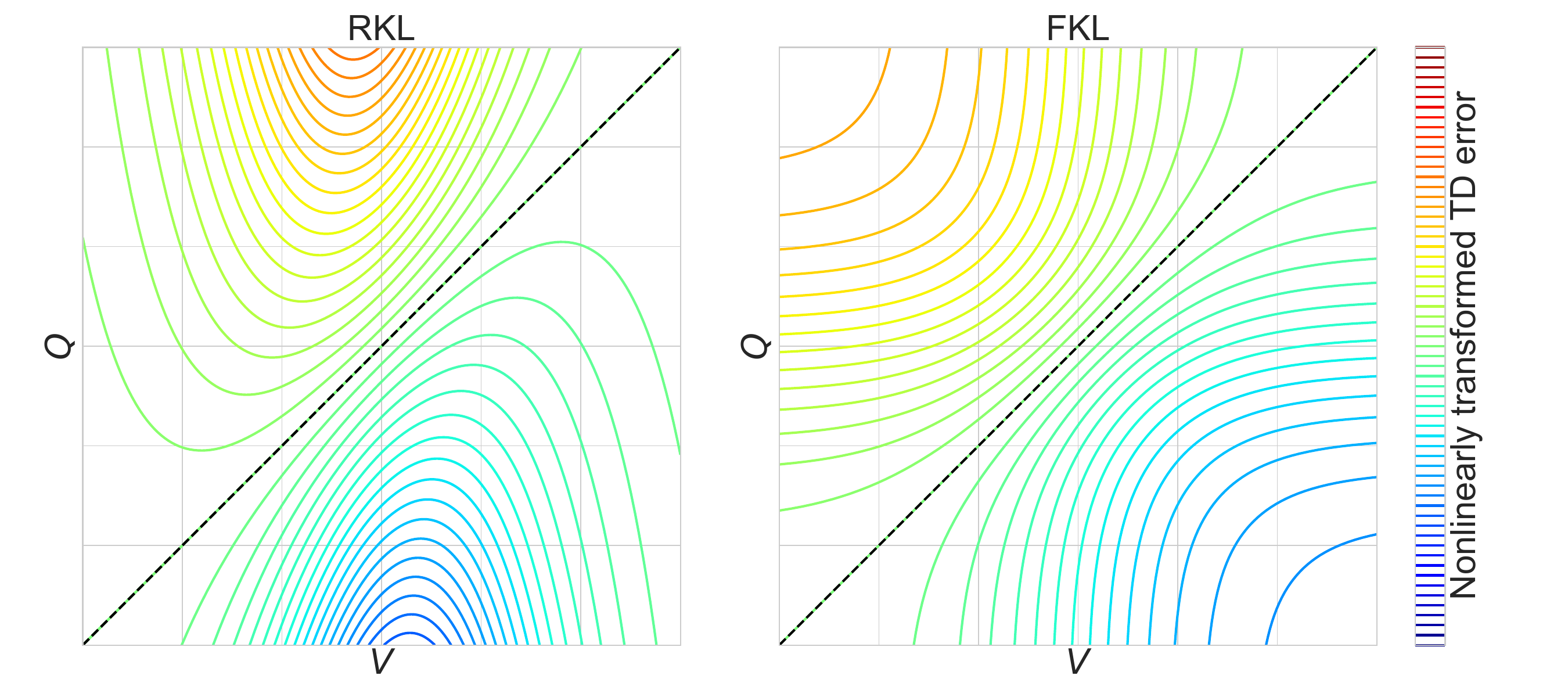}
    \caption{Analysis of nonlinear TD errors derived with forward and reverse KL divergences:
    in both cases, the gradients are vanished if the value estimate is closer to upper or lower bound;
    with reverse KL divergence (left), the contour lines are mirrored for the dashed line ($\delta=0$);
    with forward KL divergence (right), the contour lines are symmetric for the dashed line.
    }
    \label{fig:tderr_saturation}
\end{figure}

First, let's find the common point in the two nonlinear TD errors.
In Fig.~\ref{fig:tderr_saturation}, the contour lines for $\delta^\mathrm{RKL,FKL}$ are depicted.
Note that the conventional method has the contour lines parallel to the dashed line drawn diagonally.
Although they have distinct nonlinearities as the upper and lower bounds get closer, they appear to have similar shapes in the vicinity of the dashed lines, namely when $|\delta| \ll 1$ holds.

Specifically, $|\delta| \ll 1$ is assumed for applying the first-order approximation to $\delta^\mathrm{FKL}$.
\begin{align}
    \delta^\mathrm{FKL}(\delta) &= \sigma_{\delta + V} - \sigma_V
    \nonumber \\
    &\stackrel{|\delta| \ll 1}{\simeq} \frac{d \sigma_V}{dV} \delta
    \nonumber \\
    &= \lambda_O \sigma_V \bar{\sigma}_V \delta
    \nonumber \\
    &= \delta^\mathrm{RKL}(\delta)
\end{align}
Thus, when the TD error is sufficiently small, the two nonlinearizations are approximately equal.
In addition, unless the value estimate is close to the upper or lower bound, the nonlinearity by $\sigma_V \bar{\sigma}_V$ becomes nearly constant, making the TD error (or weight) linear as like the conventional method.

\begin{figure}[tb]
    \centering
    \includegraphics[keepaspectratio=true,width=0.96\linewidth]{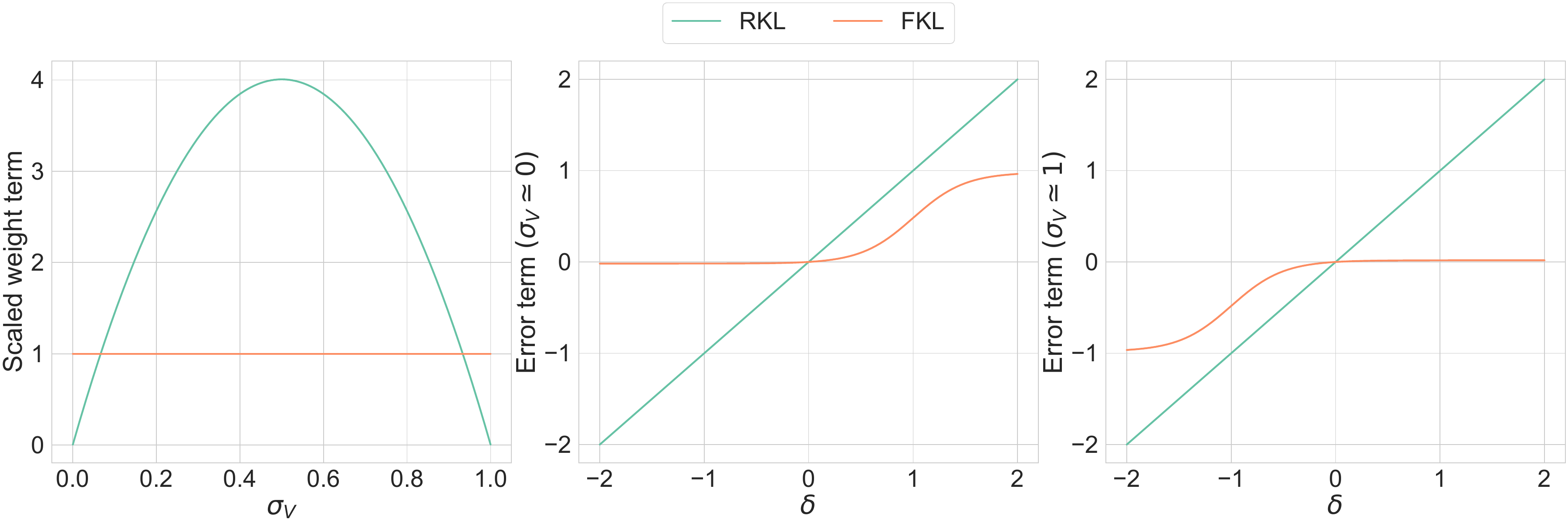}
    \caption{Decomposition of nonlinear TD errors into weight and error terms:
    the weight term (left) is saturated (or vanished) only in the case with reverse KL divergence;
    when $\sigma_V \simeq \{0, 1\}$, the error term in the case with forward KL divergence outputs almost positive or negative values.
    }
    \label{fig:tderr_saturation2}
\end{figure}

Next, the distinct points in them are analyzed.
We can find from Fig.~\ref{fig:tderr_saturation} that $\delta^\mathrm{FKL}$ has the symmetric contour lines for the dashed line, while $\delta^\mathrm{RKL}$ has the mirrored ones.
This difference is due to the influence of terms of the second order or higher, which were ignored in the approximation above, but more detailed characteristics of the respective gradients are analyzed using Fig.~\ref{fig:tderr_saturation2}.
Here, the nonlinearly transformed TD errors are decomposed into two terms: a (scaled) weight term depending on $\sigma_V$, $w(\sigma_V)$; and an error term depending on $\delta$, $e(\delta)$.
\begin{align}
    \delta^* = w^*(\sigma_V)e^*(\delta)
\end{align}
The left of Fig.~\ref{fig:tderr_saturation2} shows the nonlinearity by the weight term, the center and right are the nonlinearity by the error term.
As the definitions indicate, $\delta^\mathrm{RKL}$ has the nonlinear weight term while $\delta^\mathrm{FKL}$ has the constant.
On the other hand, $\delta^\mathrm{RKL}$ is linear while $\delta^\mathrm{FKL}$ saturates the error term.
For example, when $\sigma_V \simeq \{0, 1\}$, $w^\mathrm{RKL}$ approaches zero, causing the gradients to vanish almost completely; however, in $\delta^\mathrm{FKL}$, the positive or negative gradients remain.
Otherwise, if $\sigma_V \simeq 0.5$, the gradients with $\delta^\mathrm{RKL}$ can grow almost without limit in proportion to $\delta$, whereas in $\delta^\mathrm{FKL}$, the weight of the gradients is limited within $\pm0.5$.
Thus, while both nonlinearities function to implicitly exclude target data from learning by setting the gradients to zero, it is clear that the conditions under which they are activated and their side effects differ.

\section{Further developments}
\label{sec:develop}

\subsection{Extension to pseudo-quantization}

Based on the above analysis results, the derived gradients are vanished due to saturation when the value function exceeds its estimated upper or lower bound, $R_u, R_l$, implicitly excluding that data from learning.
However, this might be activated only for extreme estimation errors and/or reward noise.
It might help stabilize learning if this implicit data exclusion is activated in more situations.
Therefore, noting that saturation corresponds to binary quantization, the optimality probability is extended for (pseudo) multi-level quantization.

\begin{figure}[tb]
    \centering
    \includegraphics[keepaspectratio=true,width=0.96\linewidth]{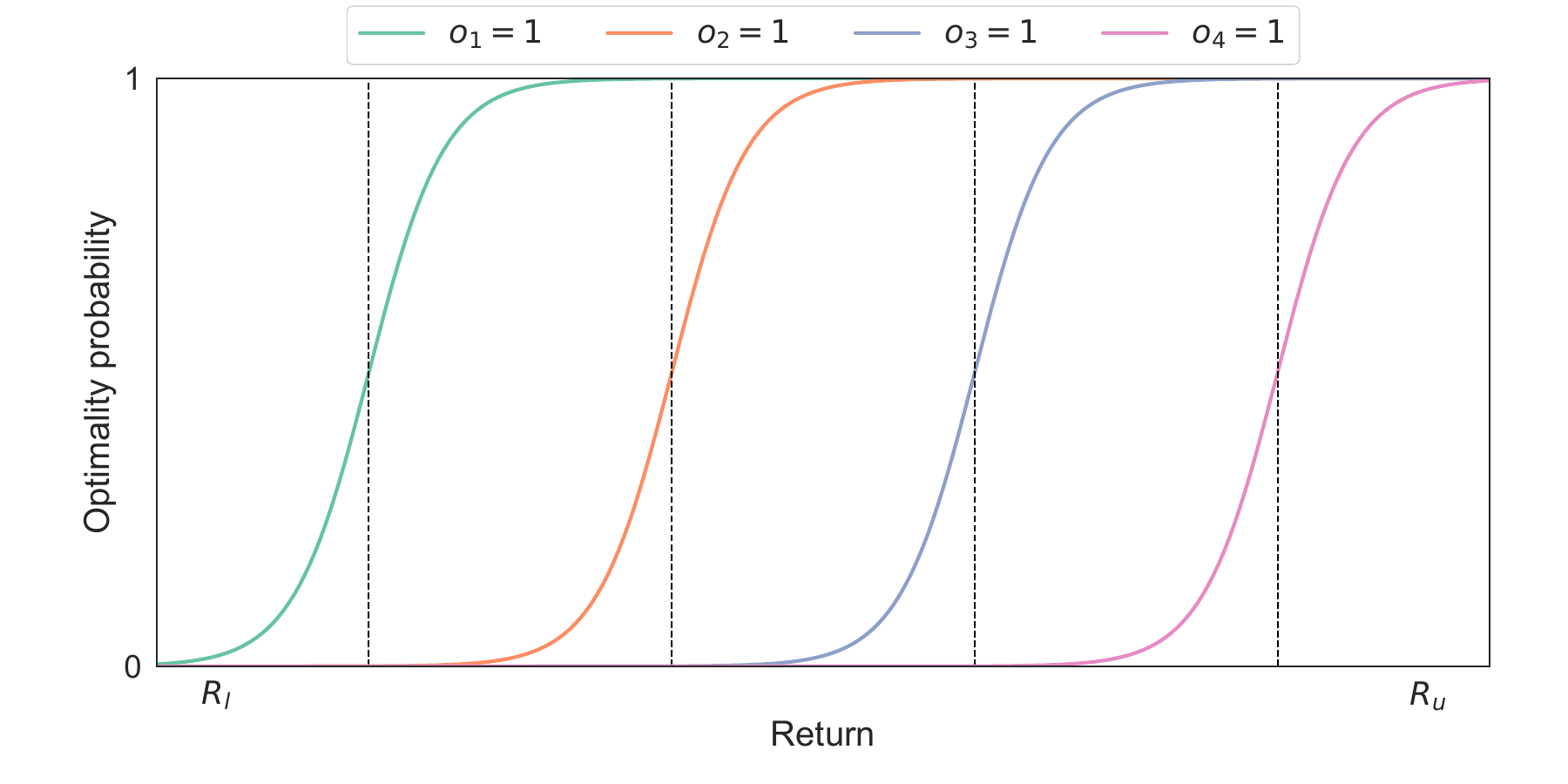}
    \caption{Extension of the model for optimality probability:
    since sigmoid function always makes its output within $[0, 1]$, multiple optimality variables can be easily introduced;
    by evenly spacing them, the saturation property is expected between them as a kind of (pseudo-)quantization.
    }
    \label{fig:optimality_quantization}
\end{figure}

Specifically, the optimality probability is extendedly defined as follows:
\begin{align}
    p(O=1) = \prod_{l=1}^L p(O_l=1)
\end{align}
where $L \in \mathbb{N}$ relates to the level of quantization.
If $L=1$, this is consistent with the original definition; and if $L>1$, this evaluates the optimality $L+1$ levels.
This definition violates the probability mass function requirement when modeling with exponential function.
Setting a common utopia point to avoid this issue loses the meaning of the extension.
In contrast, the definition introduced in this paper using sigmoid function always satisfies $[0, 1]$ regardless of the specified $\mu_O$ and $\lambda_O$, thus allowing this extension without issue.
Furthermore, if the model for $O_l$ is specified with appropriate $\mu_{O_l}$ and $\lambda_{O_l}$, $O_{<l}=1$ can also be expected when $O_l=1$ is expected.
\begin{align}
    \mu_{O_l} &= R_l + \frac{R_u - R_l}{L+1} l
    \\
    \lambda_O &= \frac{\lambda(L+1)}{R_u - R_l}
\end{align}
The case with $L=4$ is illustrated in Fig.~\ref{fig:optimality_quantization}.

As this extension yields a kind of multivariate independent distribution, the total KL divergence (and other most divergences) can be replaced to the sum of KL divergence for each component.
\begin{align}
    \mathrm{KL}(p \mid q) = \sum_{l=1}^L \mathrm{KL}(p_l \mid q_l)
\end{align}
In other words, the value function can be optimized by computing the gradients derived from the previous optimization problem in eq.~\eqref{eq:problem_val} with $\mu_{O_l}$ and $\lambda_O$ and summing them together.
As a result, the value function learns from data near $\mu_{O_l}$ and excludes data near the boundaries.
This means it behaves robustly against estimation errors and reward noise, whose scale is close to $\lambda/\lambda_O$, while still excluding outliers exceeding $R_u$ or $R_l$ as before.

On the other hand, the policy optimization problem basically requires modification.
Specifically, the conditional policies $\pi(a \mid s; O_l)$ on $O_l$ with $l=1,\ldots,L$ are introduced.
Then, $\pi(a \mid s; O)$ is replaced in the optimization problem in eq.~\eqref{eq:problem_pol} with $\pi(a \mid s; O_l)$.
Their respective gradients are summed finally.
This enables the same processing as that for the value function, and the policy is gradually optimized by switching the optimization target from $\pi(a \mid s; O_1)$ to $\pi(a \mid s; O_L)$.

Anyway, the overall gradients are generally described as follows:
\begin{align}
    g_\theta^* &\propto \mathbb{E}_{p_e, b}\left[- \nabla_\theta V(s) \left\{L^{-1} \sum_{l=1}^L \delta^*(\delta; \mu_{O_l}, \lambda_O)\right\}\right]
    \label{eq:grad_val_multi} \\
    g_\phi^* &\propto \mathbb{E}_{p_e,\pi \text{ or }b}\left[- \nabla_\phi \ln \pi(a \mid s) \left\{L^{-1} \sum_{l=1}^L \delta^*(\delta; \mu_{O_l}, \lambda_O)\right\}\right]
    \label{eq:grad_pol_multi}
\end{align}
Here, $L^{-1}$ is added for scaling.

\subsection{Fusion of forward/reverse KL divergences}

Analytical results indicate that the gradients derived using the forward or reverse KL divergence exhibit distinct nonlinear characteristics, suggesting they may have specific strengths and weaknesses.
To exploit their strengths together, a straightforward approach would be to adopt the following Jeffreys divergence.
\begin{align}
    \mathrm{Jef}(p \mid q) = \mathrm{KL}(p \mid q) + \mathrm{KL}(q \mid p)
\end{align}
In this case, we can simply add the gradients derived from both sides.
However, this approach is likely to prevent one side from utilizing the other's inherent strengths, as the other side inhibits the function that eliminates the gradients held by the first.

Hence, as a more careful fusion of both gradients, the following JS divergence is considered.
\begin{align}
    \mathrm{JS}(p \mid q) &= \frac{1}{2} \left\{\mathrm{KL}(p \mid m) + \mathrm{KL}(q \mid m) \right\}
\end{align}
where $m = (p + q) / 2$.

Although this JS divergence is required to pass to eqs.~\eqref{eq:problem_val} and~\eqref{eq:problem_pol} for analytically computing the gradients, the derivation becomes too complex compared to the cases with forward/reverse KL divergences.
Therefore, the gradient for one of the probabilities input into JS divergence is first computed as it is helpful to simplify the derivation.
Specifically, given $p_\zeta$ as the target probability with $\zeta$ the parameter, the gradient can be computed as follows:
\begin{align}
    \nabla_\zeta \mathrm{JS}(p_\zeta \mid q) &= \frac{1}{2} \mathbb{E}_{p_\zeta}[\nabla_\zeta \ln p_\zeta (\ln p_\zeta - \ln (p_\zeta + q) + \ln 2)]
    \nonumber \\
    &+ \frac{1}{2} \mathbb{E}_{p_\zeta}[\nabla_\zeta \ln p_\zeta - p_\zeta / (p_\zeta + q) \nabla_\zeta \ln p_\zeta]
    \nonumber \\
    &+ \frac{1}{2} \mathbb{E}_{q}[ - p_\zeta / (p_\zeta + q) \nabla_\zeta \ln p_\zeta]
    \nonumber \\
    &= \frac{1}{2} \mathbb{E}_{p_\zeta}[ \nabla_\zeta \ln p_\zeta (\ln D + 1 - D) ]
    \nonumber \\
    &+ \frac{1}{2} \mathbb{E}_{p_\zeta}[ - q / (p_\zeta + q) \nabla_\zeta \ln p_\zeta]
    \nonumber \\
    &= \frac{1}{2} \mathbb{E}_{p_\zeta}[ \nabla_\zeta \ln p_\zeta \ln D ] \propto \mathbb{E}_{p_\zeta}[ \nabla_\zeta \ln p_\zeta \ln D ]
\end{align}
where $D = p_\zeta / (p_\zeta + q)$ (and $1 - D = q / (p_\zeta + q)$ accordingly).
Note that even when the input order of $p_\zeta$ and $q$ is swapped, the gradient remains the same since JS divergence is symmetric.
In addition, $1/2$ was excluded as it is a constant.

With this, the gradient for the value function can be easily derived by $p_\zeta = p(O \mid s)$, $\zeta=\theta$, and $q = p(O \mid s,a)$.
\begin{align}
    g_\theta^\mathrm{JS} &= \mathbb{E}_{p_e, b}[\mathbb{E}_{p(O \mid s)}[\nabla_\theta \ln p(O \mid s) \ln D]]
    \nonumber \\
    &= \mathbb{E}_{p_e, b}[\sigma_v / \sigma_V \nabla_\theta V(s) \lambda_O \sigma_V \bar{\sigma}_V (\ln \sigma_V - \ln (\sigma_V + \sigma_Q))
    \nonumber \\
    &\quad\quad\quad\quad - \bar{\sigma}_V /\bar{\sigma}_V \nabla_\theta V(s) \lambda_O \sigma_V \bar{\sigma}_V (\ln \bar{\sigma}_V - \ln (\bar{\sigma}_V + \bar{\sigma}_Q))]
    \nonumber \\
    &= \mathbb{E}_{p_e, b}\left[ \nabla_\theta V(s) \lambda_O \sigma_V \bar{\sigma}_V \left( \ln \frac{\sigma_V}{\sigma_V + \sigma_Q} - \ln\frac{\bar{\sigma}_V}{\bar{\sigma}_V + \bar{\sigma}_Q} \right) \right]
    \label{eq:grad_val_js_raw}
\end{align}

Similarly, the gradient for the policy is also derived by $p_\zeta = \pi(a \mid s)$, $\zeta=\phi$, and $q = \pi(a \mid s; O=\{0,1\})$.
\begin{align}
    g_\phi^\mathrm{JS} &= \mathbb{E}_{p_e,\pi}[\nabla_\phi \ln \pi(a \mid s) \ln D_{O=1}] - \mathbb{E}_{p_e,\pi}[\nabla_\phi \ln \pi(a \mid s) \ln D_{O=0}]
    \nonumber \\
    &= \mathbb{E}_{p_e,\pi}\left[ \nabla_\phi \ln \pi(a \mid s) \ln \frac{\pi(a \mid s) + \pi(a \mid s; O=0)}{\pi(a \mid s) + \pi(a \mid s; O=1)} \right]
    \nonumber \\
    &= \mathbb{E}_{p_e,\pi}\left[ \nabla_\phi \ln \pi(a \mid s) \ln \frac{\sigma_V \{\bar{\sigma}_V\pi(a \mid s) + \bar{\sigma}_Q b(a \mid s)\}}{\bar{\sigma}_V\{\sigma_V\pi(a \mid s) + \sigma_Q b(a \mid s)\}} \right]
    \nonumber \\
    &\stackrel{b \simeq \pi}{\simeq} \mathbb{E}_{p_e,\pi}\left[ \nabla_\phi \ln \pi(a \mid s) \left( \ln \frac{\sigma_V}{\sigma_V + \sigma_Q} - \ln\frac{\bar{\sigma}_V}{\bar{\sigma}_V + \bar{\sigma}_Q} \right) \right]
    \label{eq:grad_pol_js_raw}
\end{align}
Here, $b \simeq \pi$ can be satisfied with the regularization method like~\citep{kobayashi2024revisiting}.

As a result, $g_\theta^\mathrm{JS}$ and $g_\phi^\mathrm{JS}$ have a similar structure, but the common weight term would be numerically unstable.
In addition, the similarity between this structure and the others previously obtained is difficult to discern.
Therefore, that term is reformulated as follows:
\begin{align}
    & \ln \frac{\sigma_V}{\sigma_V + \sigma_Q} - \ln \frac{\bar{\sigma}_V}{\bar{\sigma}_V + \bar{\sigma}_Q}
    \nonumber \\
    =& - \ln \frac{\sigma_V + \sigma_Q}{\sigma_V} + \ln \frac{\bar{\sigma}_V + \bar{\sigma}_Q}{\bar{\sigma}_V}
    \nonumber \\
    =& - \ln \{1 + \exp(\ln\sigma_Q - \ln \sigma_V)\} + \ln \{1 + \exp (\ln \bar{\sigma}_Q - \ln \bar{\sigma}_V)\}
    \nonumber \\
    =& - \mathrm{sp}(\ln\sigma_Q - \ln \sigma_V) + \mathrm{sp}(\ln \bar{\sigma}_Q - \ln \bar{\sigma}_V)
    \nonumber \\
    =& - \mathrm{sp}\{- \mathrm{sp}(-\lambda_O(Q(s,a) - \mu_O)) + \mathrm{sp}(-\lambda_O(V(s) - \mu_O))\}
    \nonumber \\
    &+ \mathrm{sp}\{- \mathrm{sp}(\lambda_O(Q(s,a) - \mu_O)) + \mathrm{sp}(\lambda_O(V(s) - \mu_O))\}
    \nonumber \\
    =& - \mathrm{sp}\{\lambda_O (Q - V) - \mathrm{sp}(\lambda_O(Q(s,a) - \mu_O)) + \mathrm{sp}(\lambda_O(V(s) - \mu_O))\}
    \nonumber \\
    &+ \mathrm{sp}\{\mathrm{sp}(\lambda_O(Q(s,a) - \mu_O)) - \mathrm{sp}(\lambda_O(V(s) - \mu_O))\}
    \nonumber \\
    &- \mathrm{sp}(\lambda_O(Q(s,a) - \mu_O)) + \mathrm{sp}(\lambda_O(V(s) - \mu_O))
    \nonumber \\
    =& \mathrm{sp}(\lambda_O (Q - V) - \Delta) + \mathrm{sp}(\Delta) - \Delta
    \nonumber \\
    =& \Delta - \lambda_O (Q - V) - \mathrm{sp}(\Delta - \lambda_O (Q - V)) + \mathrm{sp}(\Delta) - \Delta
    \nonumber \\
    =& - \lambda_O [(Q - V) + \lambda_O^{-1} \{\mathrm{sp}(\Delta - \lambda_O (Q - V)) - \mathrm{sp}(\Delta)\}]
    \nonumber \\
    \simeq& - \lambda_O [\delta + \lambda_O^{-1} \{\mathrm{sp}(\Delta - \lambda_O \delta) - \mathrm{sp}(\Delta)\}]
\end{align}
where $\Delta = \mathrm{sp}(\lambda_O(Q(s,a) - \mu_O)) - \mathrm{sp}(\lambda_O(V(s) - \mu_O))$.
As the softplus function has a numerically stable implementation in the major deep learning libraries, this form enables stable computation even when the value function approaches its lower or upper bound.
In addition, by separating the linear and nonlinear terms for the TD error, the relationship with the case using reverse KL divergence is revealed.

Substituting the obtained term into $g_{\theta,\phi}^\mathrm{JS}$ yields the (numerically stable) raw gradients.
Similar to the cases with forward and reverse KL divergences, they should be aligned so that they can be regarded to be weighted by a common nonlinearly transformed TD error.
Here, note that during the above analysis, the nonlinear component was split into a coefficient term for the TD error and a transformation term for the TD error.
The case with reverse KL divergence possessed the former, while the case with forward KL divergence possessed the latter.
We see that the gradients derived here possess both nonlinear components, but the former matches the case with reverse KL divergence.
Returning to the purpose of examining JS divergence, (namely, the appropriate fusion to exert the strengths of both), this approach risks inheriting the characteristics of the case with reverse KL divergence excessively compared to that of the case with forward KL divergence.
Indeed, as subsequent analysis will show, applying the same alignment as before (i.e. multiplying the value function gradient by $1/\lambda_O$ and the policy gradient by $\sigma_V\bar{\sigma}_V$) results in a small gradient scale, making it impossible to achieve the desired behavior.

Therefore, to obtain intermediate characteristics between the two cases, the following aligned gradients, where the gradient of the value function is multiplied by $1/(\lambda_O \sqrt{\sigma_V\bar{\sigma}_V})$ and the gradient of the policy is multiplied by $\sqrt{\sigma_V\bar{\sigma}_V}$, are proposed.
\begin{align}
    g_\theta^\mathrm{JS} &\propto \mathbb{E}_{p_e, b}[- \nabla_\theta V(s) \lambda_O \sqrt{\sigma_V\bar{\sigma}_V}
    \label{eq:grad_val_js} \\
    & \quad\quad\quad\quad [(Q(s,a) - V(s)) + \lambda_O^{-1} \{\mathrm{sp}(\Delta - \lambda_O (Q(s,a) - V(s))) - \mathrm{sp}(\Delta)\}]]
    \nonumber \\
    g_\phi^\mathrm{JS} &\propto \mathbb{E}_{p_e,\pi}[- \nabla_\phi \ln \pi(a \mid s) \lambda_O \sqrt{\sigma_V\bar{\sigma}_V}
    \label{eq:grad_pol_js} \\
    & \quad\quad\quad\quad [(Q(s,a) - V(s)) + \lambda_O^{-1} \{\mathrm{sp}(\Delta - \lambda_O (Q(s,a) - V(s))) - \mathrm{sp}(\Delta)\}]]
    \nonumber
\end{align}
Here, the nonlinear TD error can be defined as follows:
\begin{align}
    \delta^\mathrm{JS}(\delta) = \lambda_O \sqrt{\sigma_V\bar{\sigma}_V} [\delta + \lambda_O^{-1} \{\mathrm{sp}(\Delta - \lambda_O \delta) - \mathrm{sp}(\Delta)\}]
    \label{eq:td_err_js}
\end{align}

\subsection{Analysis}

\begin{figure}[tb]
    \centering
    \includegraphics[keepaspectratio=true,width=0.96\linewidth]{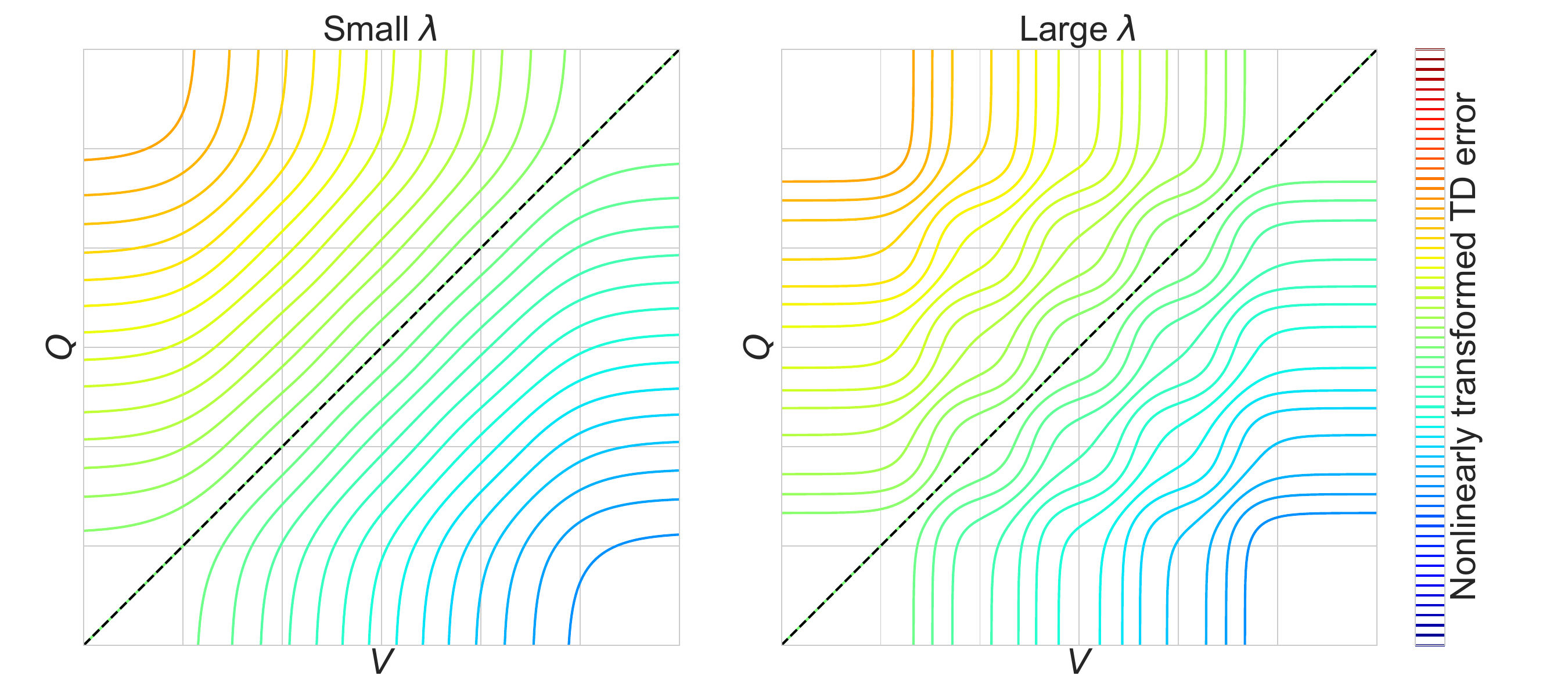}
    \caption{Analysis of nonlinear TD errors with the extension of optimality probability:
    if $\lambda$ is too small (left), the saturation property between adjacent optimalities is canceled out;
    by setting $\lambda$ is enough large (right), the expected pseudo-quantization with vanishment of gradients between adjacent optimalities is observed.
    }
    \label{fig:tderr_quantization}
\end{figure}

First, the contour lines with pseudo-quantization are visualized.
Here, using the case with forward KL divergence as an example, Fig.~\ref{fig:tderr_quantization} is depicted.
The contour lines on the left and right were plotted by varying $\lambda$ (small and large), respectively.
As can be seen from the figure, it appears that the benefits of pseudo-quantization are not found when $\lambda$ is small.
This is because, near the partitioned boundaries, the gradients regarding the adjacent $O_{l,l+1}$ are summed; thus, if $\lambda$ is small and the nonlinear TD error is not sufficiently small, the reduction can be canceled out by another nonlinear TD error.
On the other hand, in the contour lines on the right, where $\lambda$ is set sufficiently large, a wave-like shape can be observed.
This indicates that the gradients become smaller than usual near each boundary, and learning is suppressed by pseudo-quantization.
Qualitatively, since the sigmoid function converges around $\pm4$, $\lambda=4$ is natural choice to obtain the benefits of this pseudo-quantization.

\begin{figure}[tb]
    \centering
    \includegraphics[keepaspectratio=true,width=0.96\linewidth]{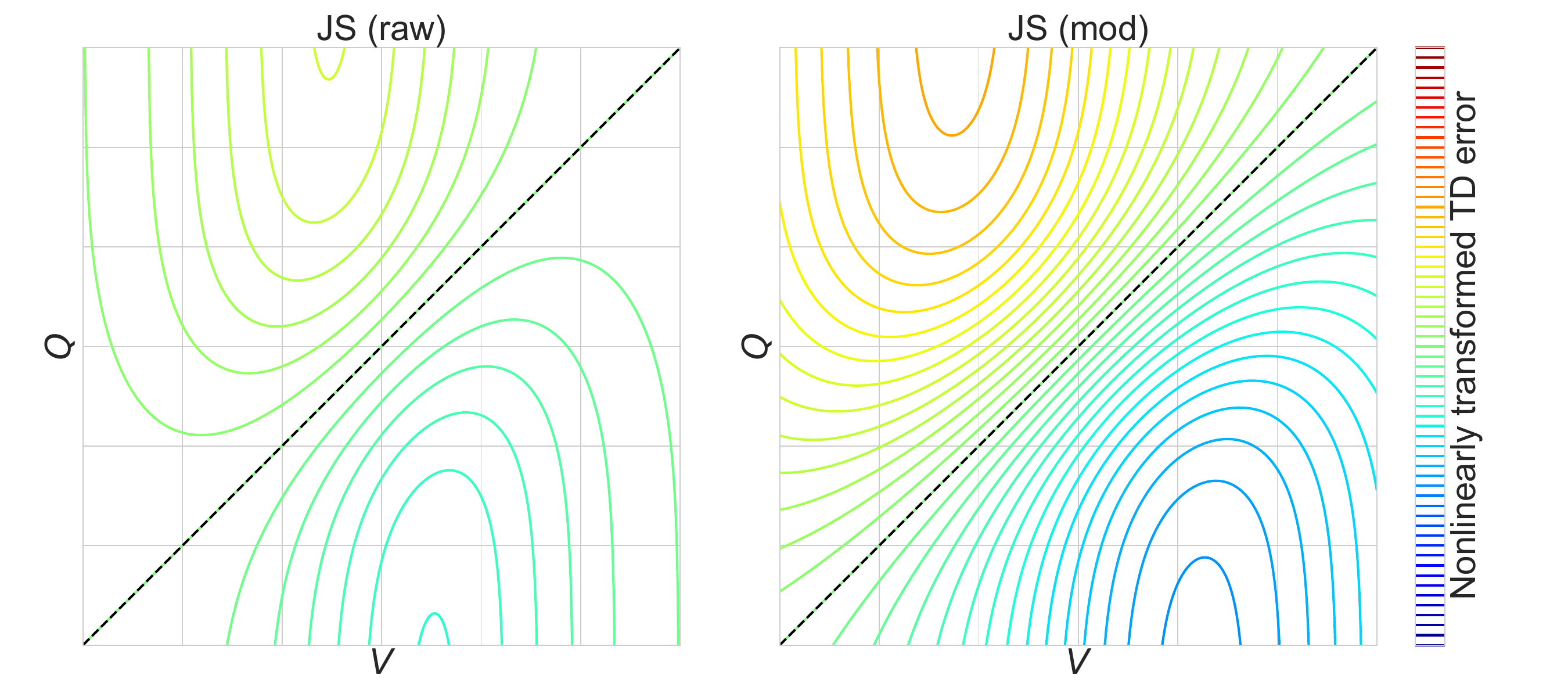}
    \caption{Analysis of nonlinear TD errors with JS divergence:
    if the alignment trick is not modified from the cases with forward and reverse KL divergences (left), the shape of contour lines is similar to the case with reverse KL divergence while widening the spaces between lines;
    with the modified alignment trick (right), the shape of contour lines are mixed up with two previous cases with the appropriate spaces.
    }
    \label{fig:tderr_saturation_js}
\end{figure}

Next, the characteristics of the gradients derived using JS divergence are qualitatively analyzed.
At the same time, the conventional alignment trick in the cases with forward and reverse KL divergences are compared to the modified one proposed in the above.
The contour lines (setting $L=1$ for simplicity) are again plotted in Fig.~\ref{fig:tderr_saturation_js}.
The left side shows the case with the conventional alignment trick, and it is immediately apparent that the shape of the contour lines is similar to that in the case with reverse KL divergence.
Furthermore, the wide spacing between the contour lines indicates that the nonlinear TD error hardly fluctuate.
In other words, because the weight term is dominant and combined with the nonlinearity of the error term, the gradient tends to vanish in all regions.
On the other hand, with the modified alignment (see the right side), the behavior in the lower-left and upper-right regions is similar to that in the case with forward KL divergence, exhibiting symmetry with respect to the dashed line.
Furthermore, the upper-left and lower-right regions exhibit a mirrored structure similar to that in the case with reverse KL divergence.
Thus, the characteristics of two types of gradients (with forward and reverse KL divergences) are preserved by the modified alignment trick.

\begin{figure}[tb]
    \centering
    \includegraphics[keepaspectratio=true,width=0.96\linewidth]{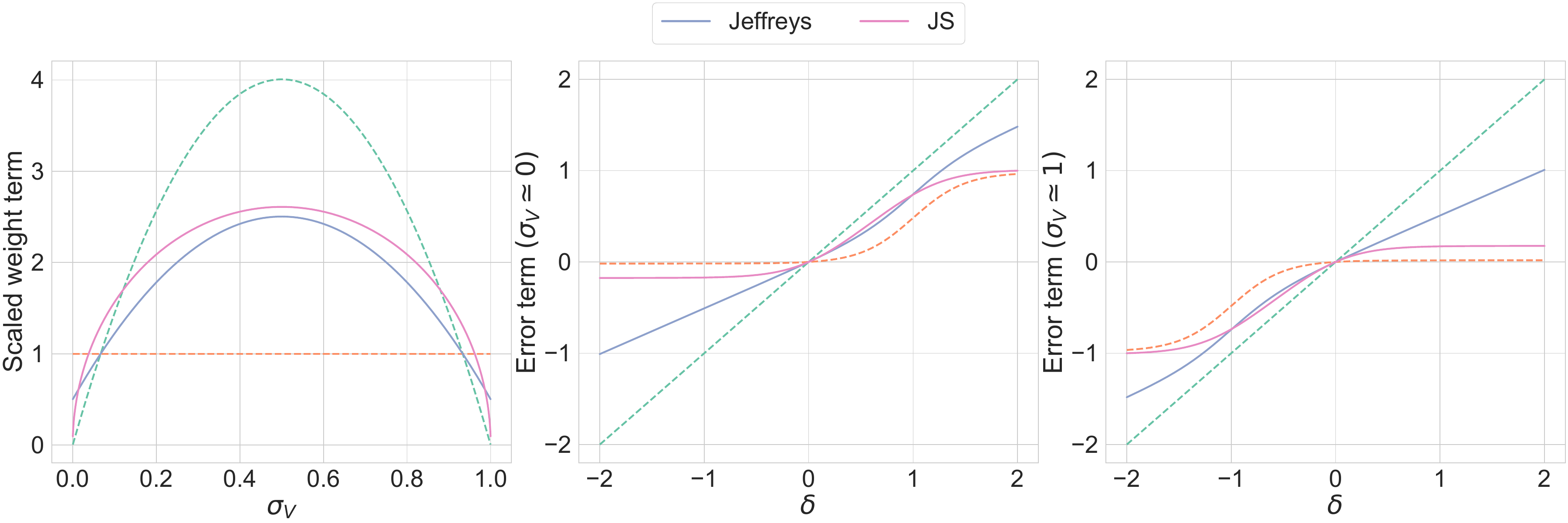}
    \caption{Weight and error terms in the cases with Jeffreys and JS divergences:
    in both terms, the case with Jeffreys divergence loses the saturation property;
    on the other hand, the case with JS divergence can inherit it in both terms while relaxing its restriction.
    }
    \label{fig:tderr_saturation_js2}
\end{figure}

Finally, the decomposed terms are plotted in Fig.~\ref{fig:tderr_saturation_js2} in order to compare the case with Jeffreys divergence (i.e. the mean of gradients derived by forward and reverse KL divergences).
Note that to make the comparison easier, the results in Fig.~\ref{fig:tderr_saturation2} are shown with dashed lines with the same colors.
With Jeffreys divergence, each term has no saturation although nonlinearities are remained.
That is, the adverse effects of value estimation errors and noisy rewards cannot be sufficiently mitigated, so the expected benefits of learning stabilization cannot be achieved from this formulation.
On the other hand, the case with JS divergence inherits saturation in both terms.
In addition, the difference in the weight term w.r.t. $\sigma_V$ becomes smaller, and the range of error term expands.
In particular, when $\sigma_V \simeq \{0, 1\}$, whereas the conventional error term using forward KL divergence could only take positive or negative values, this restriction is relaxed, and the sign of the error term is consistently inherited from the raw TD error.
Thus, the gradients with JS divergence (and the modified alignment) enables to inherit the saturation ability in both cases with forward and reverse KL divergences, while their excessive nonlinearities are relaxed.

\section{Simulations}

\subsection{Setup}

\begin{table*}[tb]
    \caption{Learning conditions}
    \label{tab:impl}
    \centering
    {\scriptsize
    \begin{tabular}{lc}
        \hline\hline
        \#Hidden layers of fully connected networks & $2$
        \\
        \#Neurons for each hidden layer & $100$
        \\
        Activation function & Squish~\citep{barron2021squareplus,kobayashi2023design}
        \\
        Normalization & RMSNorm~\citep{zhang2019root}
        \\
        Value estimation & Ensemble w/ underestimation~\citep{kobayashi2025intentionally}
        \\
        Policy distribution & Student's t-distribution~\citep{kobayashi2019student}
        \\
        \hline
        Discount factor $\gamma$ & $0.99$
        \\
        Optimizer & AdaTerm~\citep{ilboudo2023adaterm}
        \\
        Update of target networks & CAT-soft update~\citep{kobayashi2024consolidated}
        \\
        Stabilization & ERC~\citep{kobayashi2024revisiting}
        \\
        \hline
        Buffer size $|\mathcal{D}|$ & $102,400$
        \\
        Batch size $B$ & $256$
        \\
        \#Replayed data (episodic) & $|\mathcal{D}|/2$
        \\
        \hline
        Level of quantization $L+1$ & 5
        \\
        Sharpness of quantization $\lambda$ & 4.0
        \\
        \#Update for decaying $R_{u,l}$ & 200
        \\
        \hline\hline
    \end{tabular}
    }
\end{table*}

To verify the effectiveness of the proposed method, the following five simulations are conducted.
The first two experiments confirm that the derived learning rules function correctly and the effectiveness of the improvements proposed in Section~\ref{sec:develop} (i.e. pseudo-quantization and derivation based on JS divergence).
The next two experiments demonstrate the robustness of the proposed method against value estimation errors and noisy rewards.
Finally, the side effects resulting from such robustness are further explored.

The implementation of the proposed (and baseline) methods are shared in all the experiments, as summarized in Table~\ref{tab:impl}.
Note that items that cite references inherit the reference settings as-is.
Here, the estimation of $R_{u,l}$ is simply performed as follows:
\begin{align}
    \begin{split}
        R_u &\gets \beta R_u + (1 - \beta) (\max_{b\in B}V(s_b) + \epsilon)
        \\
        R_l &\gets \beta R_l + (1 - \beta) (\min_{b\in B}V(s_b) - \epsilon)
        \\
        \beta &= \epsilon^{1/K}
    \end{split}
\end{align}
where $\epsilon \ll 1$ denotes the small number for numerical stabilization (in this implementaiton, $\epsilon=10^{-5}$).
The hyperparameter $K \in \mathbb{N}$ is the number of updates required for the effect of the update value to decrease to less than $\epsilon$.
This update is performed when calculating the nonlinear TD error, that is, each time batch data $B$ is replayed from the replay buffer.
In this paper, the maximum number of replays between episodes is 200, and $K$ is set to 200 accordingly.

In all the experiments, the proposed and baseline methods train their policies 28 times using different random seeds.
Then, the tasks are performed for 100 episodes using the trained policies, and performance is evaluated based on the statistics of those returns.
Basically, the interquartile mean of 100 returns is employed to evaluate each random seed, then its box plot is visualized.
The tasks consist basically of modified versions in Gymnasium: HalfCheetah, which has restricted output; and Pusher, which has a larger and heavier target.
The former requires larger actions, so if learning is overly restricted, it cannot achieve sufficient performance.
The latter makes it difficult to redo object manipulation, so performance tends to decline if the model learns incorrectly.
Note that the last two experiments use different suitable tasks.

\subsection{Importance of appropriate quantization level}

\begin{figure}[tb]
    \centering
    \includegraphics[keepaspectratio=true,width=0.96\linewidth]{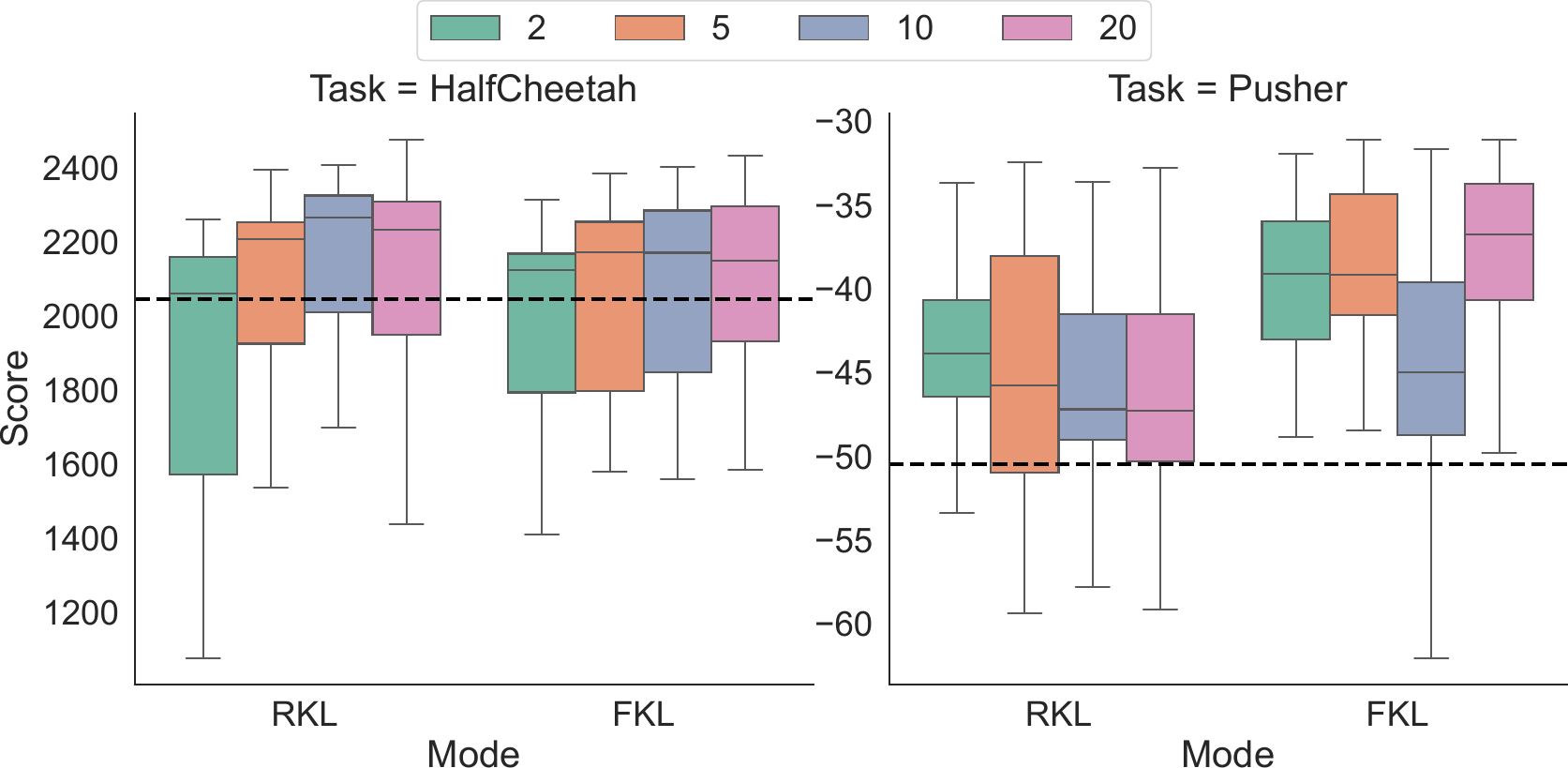}
    \caption{Results for comparing quantization level $L+1$:
    if $L+1=2$ (i.e. only saturation near $R_{u,l}$), the scores were not improved enough in all cases;
    with too large $L+1$ (i.e. $10,20$), the scores were slightly improved in HalfCheetah or decreased in Pusher;
    taking the computational cost into account as well, $L+1=5$ seems to be the best balance.
    }
    \label{fig:result_quantization}
\end{figure}

First, the value of extending to pseudo-quantization, i.e. $L>1$, is investigated.
Specifically, $L+1 = \{2, 5, 10, 20\}$ are compared.
Note that, to account for differences in the impact of divergence used to calculate gradients, the results using both forward and KL divergences are shown.

The results are summarized in Fig.~\ref{fig:result_quantization}.
The dashed lines represent the average score of the baseline method, the standard actor-critic algorithm (identical to the proposed method except for its learning rules being linear w.r.t. the TD error).
When the quantization level is $L+1=2$, where saturation occurs only around $R_{u,l}$ prior to expansion, the scores for both task divergences were lower than those of the other cases.
By setting $L>1$, a steady improvement in performance can be observed.
However, even when $L+1$ is increased to $10$ or $20$, the performance improvement was small, and it was also confirmed that scores actually decrease for the Pusher task.
Since increasing $L$ also increases computational cost, $L+1=5$ is adopted in this paper.

It is worth noting that the reason humans often adopt a five-point rating scale is similar.
That is, the benefits of finer granularity begin to diminish around the five-point mark~\citep{lissitz1975effect}.
This similarity suggests that there is room for interesting analysis to explain the relationship between the proposed method and human's decision-making.
However, that is omitted due to outside the scope of this paper.

\subsection{Inheritance of strengths from forward/reverse KL divergences}

\begin{figure}[tb]
    \centering
    \includegraphics[keepaspectratio=true,width=0.96\linewidth]{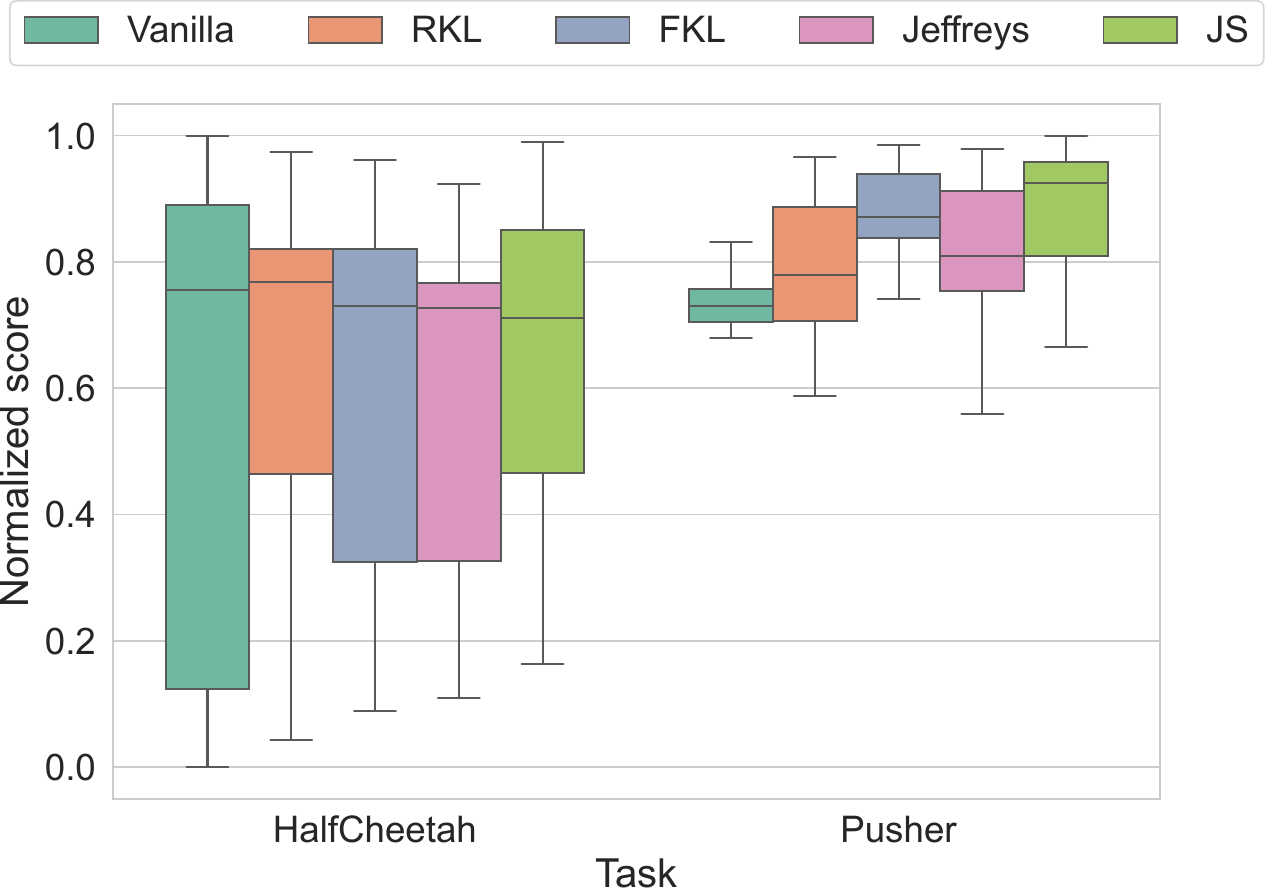}
    \caption{Results for comparing the integration way of two divergences:
    the cases with reverse and forward KL divergences improved especially HalfCheetah and Pusher, respectively;
    the case with Jeffreys divergence was closer to the worse scores in either case, while the case with JS divergence was closer to the better scores.
    }
    \label{fig:result_fusion}
\end{figure}

Next, while examining the strengths and weaknesses of the learning rules based on forward and reverse KL divergences, whether the learning rule using JS divergence effectively inherits their strengths is investigated.
Specifically, the following five conditions are compared.
\begin{itemize}
    \item Vanilla: the standard actor-critic algorithm with linear learning rules w.r.t. the TD error
    \item RKL: the learning rules derived with reverse KL divergence
    \item FKL: the learning rules derived with forward KL divergence
    \item Jeffreys: the learning rules with the mean of gradients of RKL and FKL
    \item JS: the proposed method derived with JS divergence
\end{itemize}
Note that these conditions have $L+1=5$, as described in Table~\ref{tab:impl}.

The results are summarized in Fig.~\ref{fig:result_fusion}.
First, we can see that Vanilla exhibited unstable training on HalfCheetah and did not achieve sufficient performance on Pusher.
On the other hand, RKL and FKL exceled at different tasks: RKL stabilized training on HalfCheetah, while FKL increased the success rate on Pusher.
When examining Jeffreys, it appears to inherit the weaknesses of RKL and FKL, resulting in performance closer to FKL for HalfCheetah and closer to RKL for Pusher.
In contrast, the proposed JS achieved performance closer to RKL for HalfCheetah and closer to FKL for Pusher, suggesting that it successfully inherits the strengths of both.

\subsection{Robustness to noisy TD error}

The above two points confirm that the proposed developments in Section~\ref{sec:develop} are effective.
The robustness to tasks where the TD error is noisy is then verified.
Specifically, using HalfCheetah and Pusher as base tasks, the following three points are modified individually.
\begin{itemize}
    \item NoEnsemble: the value function is predicted with a single output of model without ensembles.
    \item NonRobustTarget: the target network is updated with Polyak update, not CAT-soft update.
    \item NoisyReward: the task reward is replaced to the time-averaged reward with Gaussian noise.
\end{itemize}
The estimation of the target value tends to become unstable in NoEnsemble (due to non-reduced epistemic uncertainty) and in NonRobustTarget (due to too high update rate for Polyak update).
In NoisyReward, the raw task reward $r$ is replaced as follows:
\begin{align}
    \begin{split}
        \sigma_r^2 &\gets \frac{1}{100^{1/50}} \sigma_r^2 + \frac{1}{100^{1/50}} \left(1 - \frac{1}{100^{1/50}} \right) (\mu_r - r)^2
        \\
        \mu_r &\gets \frac{1}{100^{1/5}} \mu_r + \left(1 - \frac{1}{100^{1/5}} \right) r
        \\
        r &\sim \mathcal{N}(\mu_r, \sigma_r)
    \end{split}
\end{align}
where $\mathcal{N}(\mu_r, \sigma_r)$ denotes the diagonal normal distribution with $\mu_r$ mean and $\sigma_r$ scale.
The initial values of $\mu_r$ and $\sigma_r$ are set to be zero.

Robustness is evaluated based on the difference in return compared to the maximum return with the normal condition (as conducted above).
Since the overall statistics are more important than the performance under each of the above conditions, performance profiles~\citep{agarwal2021deep} are employed.
Specifically, all returns for each task are normalized to $[0, 1]$ using the maximum and minimum ones, and the probability that the score exceeds a given threshold within $[0, 1]$ is visualized.

For comparison, the methods excluding Jeffreys are tested.
In addition, to exhibit a certain degree of robustness, max-min entropy (MME)~\citep{han2021max}, which is a modified version of soft actor-critic (SAC)~\citep{haarnoja2018soft}, is introduced.
Since MME includes a minimization term for policy entropy, which is computed via one-sample Monte Carlo approximation, in its value estimation, the effects of value estimation errors and reward noise are unlikely to become dominant.
Furthermore, since the policy is learned by maximizing not only the value function but also its entropy, as in SAC, overfitting to incorrect value estimates can likely be avoided.
Note that the related studies on robustness discussed in Section~\ref{subsec:related} are omitted from comparison because they can be used in conjuction with the proposed method, or because the learning protocols differ significantly, making a fair comparison difficult.

\begin{figure}[tb]
    \begin{subfigure}[b]{0.48\linewidth}
        \centering
        \includegraphics[keepaspectratio=true,width=\linewidth]{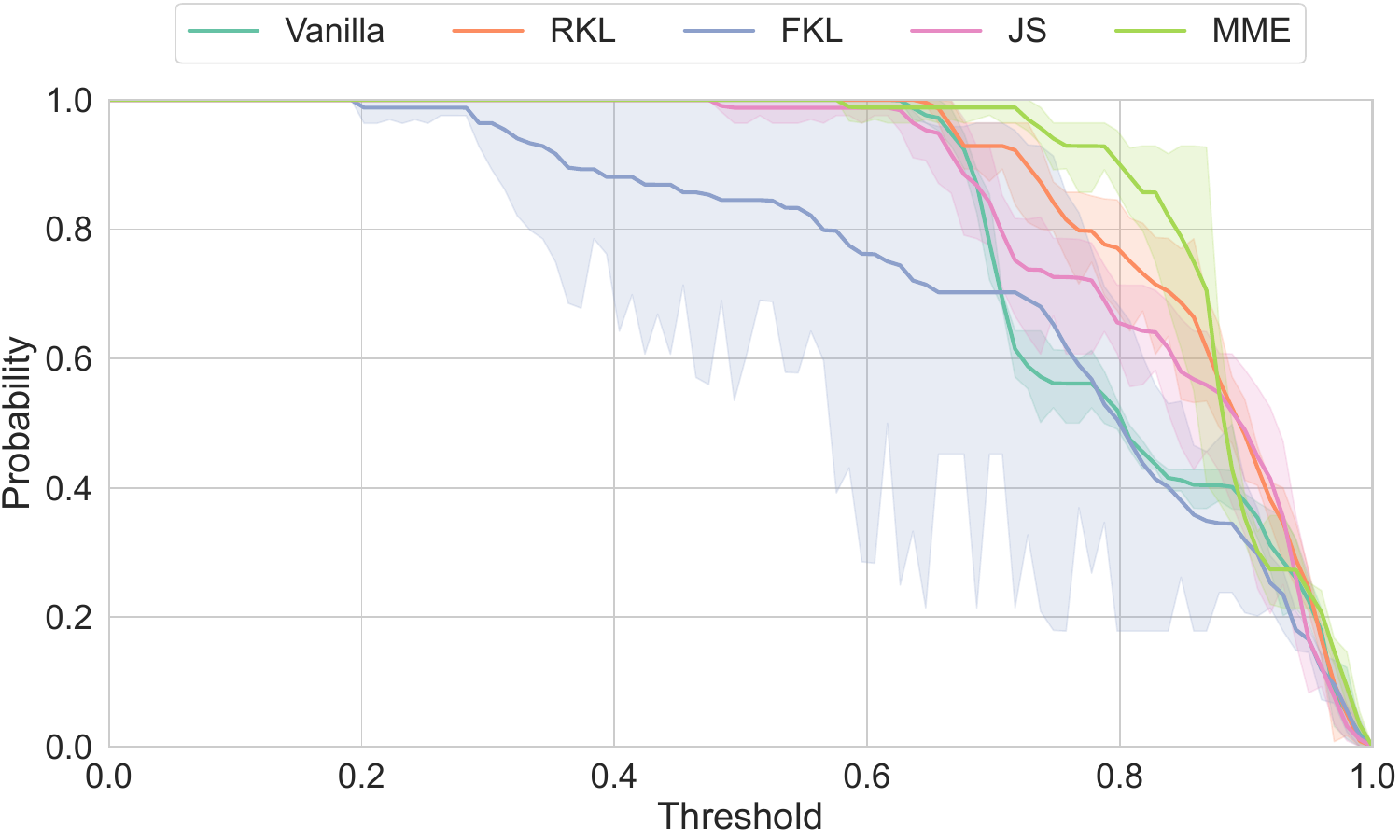}
        \subcaption{HalfCheetah}
        \label{fig:result_noisy_halfcheetah}
    \end{subfigure}
    \begin{subfigure}[b]{0.48\linewidth}
        \centering
        \includegraphics[keepaspectratio=true,width=\linewidth]{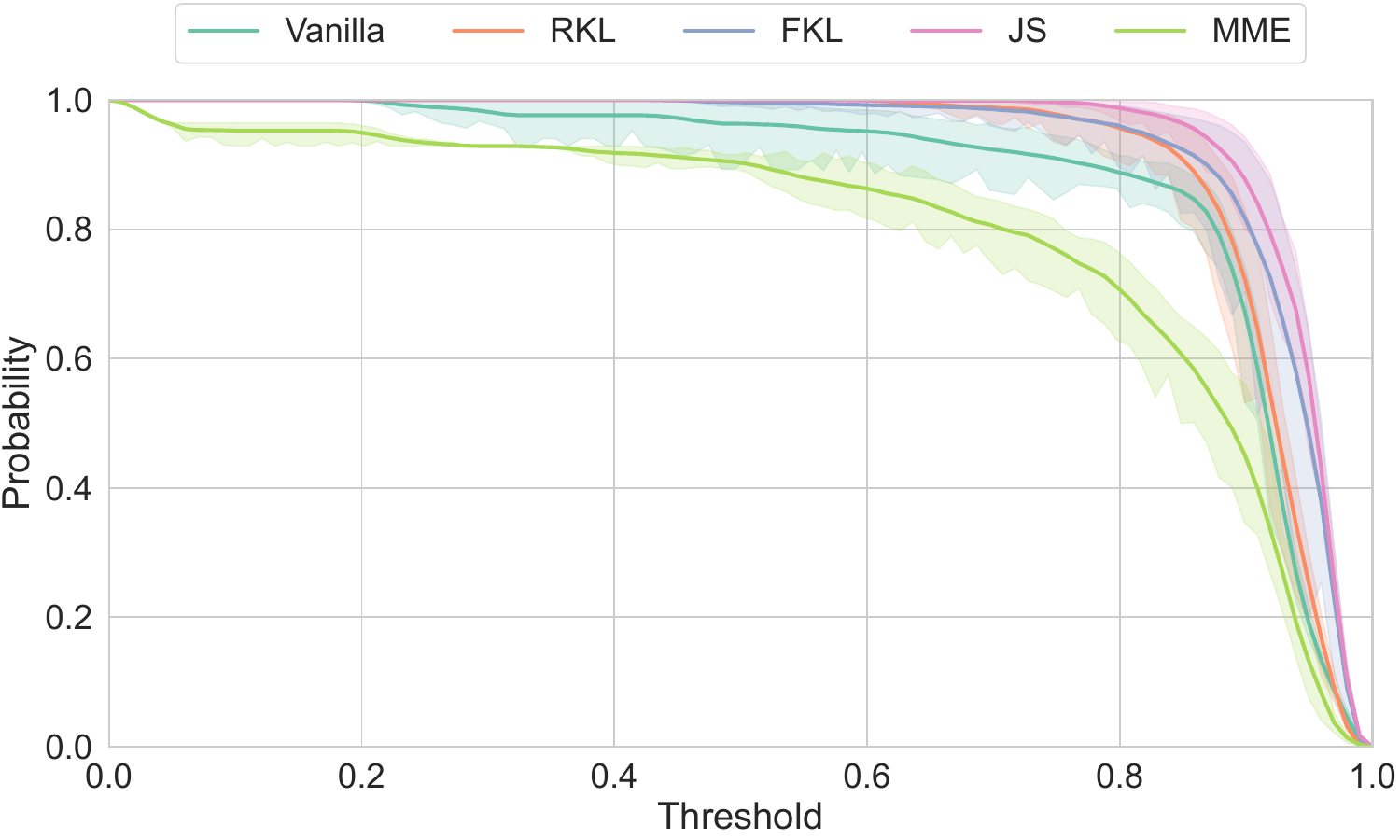}
        \subcaption{Pusher}
        \label{fig:result_noisy_pusher}
    \end{subfigure}
    \caption{Performance properties for noisy TD errors:
    the baselines (Vanilla, RKL, and FKL) had the similar tendency found in the above experiments;
    although MME achieved the best robustness in HalfCheetah, it was too fragile in Pusher;
    the proposed method (JS) achieved the stable results in both cases.
    }
    \label{fig:result_noisy}
\end{figure}

The results are depicted in Fig.~\ref{fig:result_noisy}.
First, we can see that in HalfCheetah, FKL and Vanilla exhibited low robustness with high performance degradation, whereas RKL and the proposed JS achieved roughly equivalent levels of excellent robustness.
MME appears to achieve the highest robustness, but around the threshold of $0.9$, it was outperformed by RKL and JS.
On the other hand, MME was extremely fragile in Pusher.
Vanilla and RKL were the next least performant, and FKL and JS demonstrated excellent robustness.

The results for RKL and FKL were consistent with their respective strengths and weaknesses in specific tasks, suggesting that the performance improvements in their strong tasks was due to the robustness of value estimation.
Furthermore, MME exhibited a more apparent task dependency, which is thought to be due to the policy entropy maximization term failing to suppress overfitting as much as expected.
In other words, it fell into local optima on Pusher, a task that requires more careful learning.
In contrast, the proposed JS demonstrated excellent robustness on both tasks, indicating that it effectively leverages the task-dependent robustness of both RKL and FKL.

\subsection{Robustness to guided reward}

\begin{figure}[tb]
    \centering
    \includegraphics[keepaspectratio=true,width=0.96\linewidth]{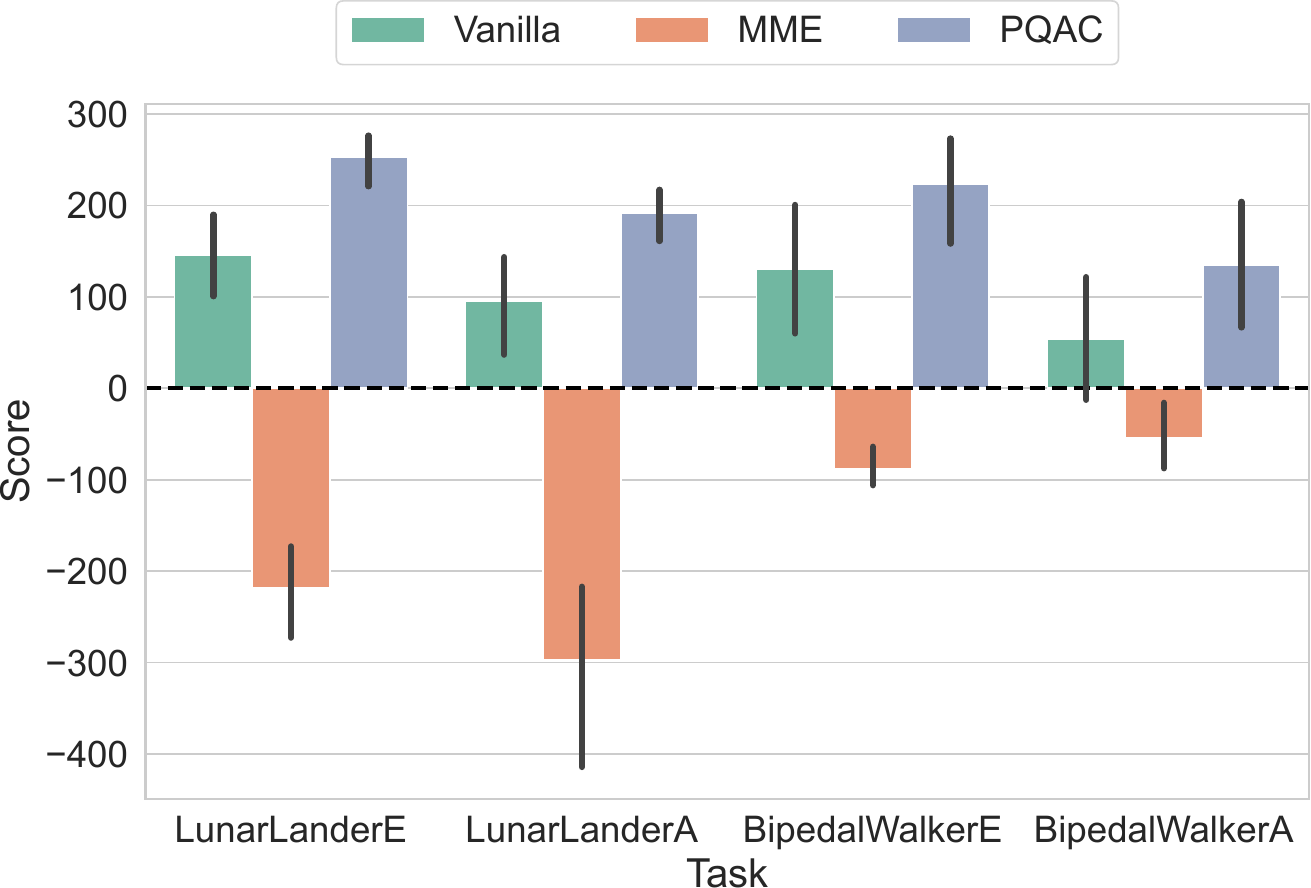}
    \caption{Results for guided RL:
    MME completely failed all tasks;
    the proposed PQAC outperformed the standard RL in all tasks, while its performance degradation from `E' (with a well-trained policy) to `A' (with a poorly trained policy) was almost equivalent to the standard RL probably due to the limitation of imitation.
    }
    \label{fig:result_guide}
\end{figure}

Furthermore, to verify robustness against noisy TD errors, additional experiments analogous to guided RL~\citep{esser2022guided} are conducted.
Specifically, expert policies $\pi^\mathrm{exp}$ trained using standard task rewards are prepared in advanced, and then, the following imitation reward (more specifically, its return) is maximized instead of the standard reward.
\begin{align}
    r = |\mathcal{A}|^{-1} \sum_{i=1}^{|\mathcal{A}|}\exp(-|a_i^\mathrm{exp} - a_i|)
\end{align}
In this case, since expert actions $a^\mathrm{exp}$ are randomly sampled from $\pi^\mathrm{exp}$, this reward is determined stochastically and acts as reward noise.
Furthermore, if the learning of $\pi^\mathrm{exp}$ is insufficient, not only this randomness increase, but also the agent may be led to different actions even in similar states, thereby destabilizing value estimation.
Furthermore, since learning becomes more difficult as the action space grows larger under this design, LunarLander (with continuous action space) and BipedalWalker in Gymnasium are solved here (also to verify their effectiveness on different tasks).

The standard learning rules (Vanilla), MME as tested above, and the proposed method with JS divergence and pseudo-quantization, so-called PQAC, are compared.
For each task, both a well-trained $\pi^\mathrm{exp}$ (suffixed with `E') and a poorly trained version (suffixed with `A') are prepared.
The results of training and testing (the return on the original task reward) are depicted in Fig.~\ref{fig:result_guide}.
As can be seen at a glance, MME was completely unable to solve the tasks.
This is probably because MME is an algorithm based on action value function $Q(s,a)$, and with the reward function described above, the gradients tend to change easily depending on the action, causing it to be strongly affected by erroneous gradients.
On the other hand, the proposed PQAC performed more stably and successfully on the tasks than Vanilla.
However, there is no significant difference between Vanilla and PQAC in the degree of performance degradation from `E' to `A'.
This and the results so far suggest that the limit of task performance achievable with the above reward is lowered, rather than indicating a lack of robustness.

\subsection{Ability to optimize in high-dimensional action spaces}

\begin{figure}[tb]
    \centering
    \includegraphics[keepaspectratio=true,width=0.96\linewidth]{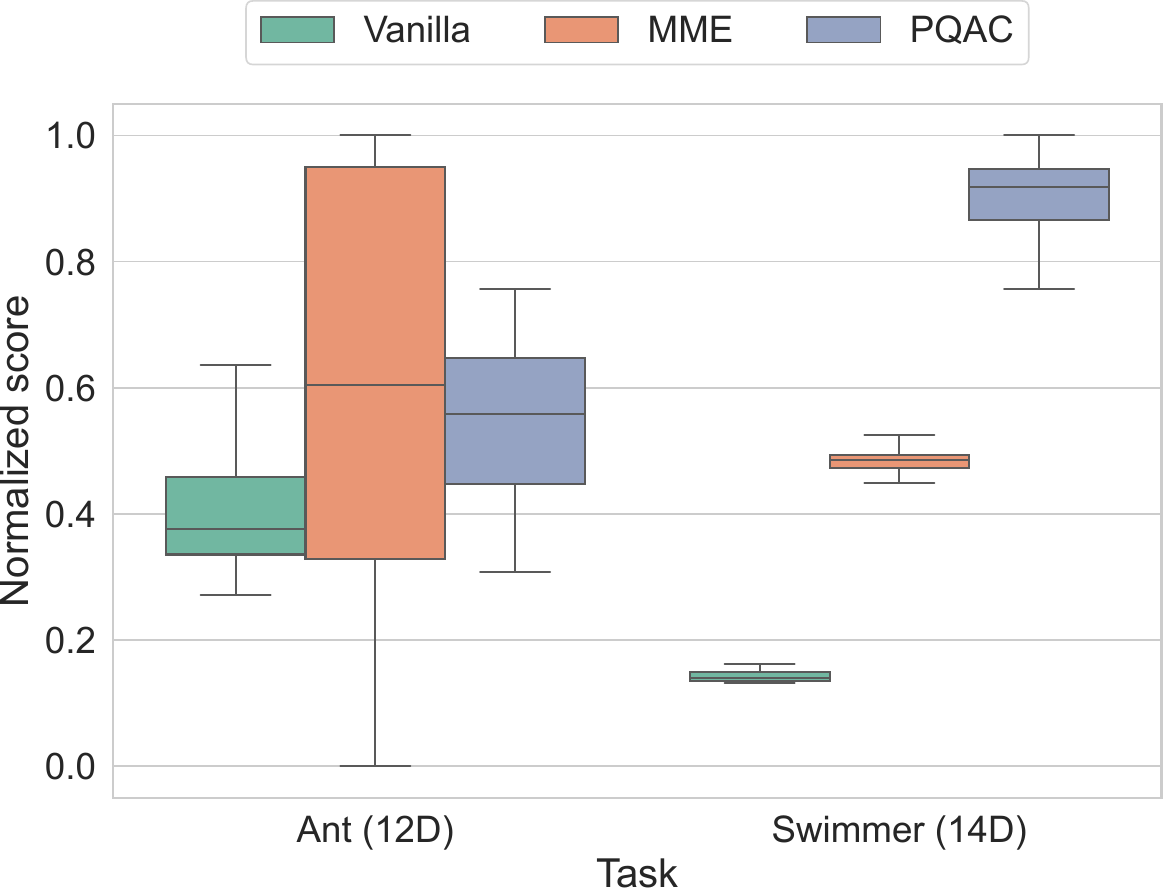}
    \caption{Results for high-dimensional action spaces:
    the standard RL could not solve these tasks;
    although MME achieved the best performance on Ant, it was unstable;
    the proposed PQAC was stable in both tasks, and achieved the best in Swimmer.
    }
    \label{fig:result_action}
\end{figure}

Finally, the side effects of the proposed method’s robustness against noisy TD errors are investigated.
Fundamentally, since the value function is the expected return for policy (and state transition probability), the greater the randomness of the policy, the more diverse the value function becomes, making it difficult to estimate~\citep{van2016deep}.
Although randomness decreases as learning progresses, the impact would persist as the size of action space increases, adversely affecting value estimation.
In other words, due to the robustness, the proposed method is expected to efficiently solve tasks with high-dimensional action spaces.

To examine it, Ant and Swimmer tasks from Gymnasium are modified to have increased number of joints.
Specifically, for Ant, three joints are given to each leg to create a 12-dimensional action space; and for Swimmer, the number of segments is increased to 15 to create a 14-dimensional action space.
The results of running these tasks using the same comparison methods as in the previous experiment are shown in Fig.~\ref{fig:result_action}.

Clearly, Vanilla did not perform well on these tasks, indicating that it struggles with high-dimensional action spaces.
Although MME achieved excellent performance in the best-case scenario for Ant, its learning became highly unstable.
As mentioned above, this instability is probably because it easily generates incorrect gradients depending on the selected actions.
In contrast, the proposed PQAC consistently outperformed Vanilla on Ant and demonstrated significantly superior performance compared to others on Swimmer.

\section{Conclusion and discussion}

This paper revisited the TD learning algorithm based on control as inference, deriving a novel algorithm capable of robust learning against noisy TD errors.
The derived learning rules for both value and policy functions were analyzed, revealing the smooth saturation of loss (or the vanishment of gradients) when the value estimate is close to its estimated uppear or lower bound.
This saturation was extended to pseudo-quantization by introducing the multi-level optimality.
In addition, to inherit different advantages in the learning rules derived with forward and rerverse KL divergences, JS divergence was approximately utilized for the novel derivation.
These benefits were verified through multiple RL benchmarks, demonstrating stable learning even when heuristics are insufficient or rewards contain noise.

As shown by the last two experiments, the robustness against noisy TD errors is expected to handle recent RL targets: i.e. preference-based reward designs~\citep{kaufmann2024survey,poddar2024personalizing,luo2024rlif,cheng2024rime} and high-dimensional robotic systems~\citep{radosavovic2024real,chen2023bi,chiappa2024acquiring}.

The former is required especially in human-in-the-loop systems involving human interaction, such as for fine-tuning foundation models~\citep{kaufmann2024survey,poddar2024personalizing} and alternatives to traditional imitation learning~\citep{luo2024rlif}.
However, since the reward itself becomes an estimate, its estimation error inevitably affects the accuracy of the value function and the noise scale of TD errors.
While methods to make this reward estimation more robust have been explored~\citep{cheng2024rime}, a synergistic effect can be expected if the RL algorithm itself becomes more robust like the proposed PQAC, allowing for a more accurate capture of human preferences.

The latter is increasingly common in the motion control of humanoid robots (with approximately 30 joints) using RL~\citep{radosavovic2024real}.
In addition, dexterous robot hands are widely under development~\citep{chen2023bi}, so loco-manipulation that utilizes all five fingers is likely to become a future target for RL.
Alternatively, body structures that actively utilize redundancy, such as musculoskeletal systems~\citep{chiappa2024acquiring}, would also be explored.
These action dimensions are close to or over 100.
Consequently, RL algorithms like PQAC, which enable robust and stable learning despite noisy TD errors than ever before, will likely be in demand.
However, there is concern that implicitly excluding some experience data from learning in such a vast exploration space could cause reduced learning efficiency; therefore, it is expected that these methods will need to be used in conjunction with active exploration techniques~\citep{sekar2020planning,russo2023model}.

In both cases, while the recent RL targets described above cannot be solved by the proposed PQAC alone, it is considered a useful foundational technology.
However, there might be room for improvements in the implementation of PQAC.
For example, a more robust and accurate estimation of $R_{u,l}$ should enable more appropriate data exclusion.
The evenly spaced $\mu_{O_l}$ might be optimized to minimize the impact of noisy TD errors.
In the near future, tricks to enhance the learning performance of PQAC will be investigated while exploring its applications described above.

\section*{Acknowledgements}

This work was supported by JSPS KAKENHI, Development and validation of a unified theory of prediction and action, Grant Number JP24H02176; and by JST CRONOS, Grant Number JPMJCS24K6.

\bibliographystyle{elsarticle-harv}
\bibliography{biblio}

\end{document}